\documentclass{article} % For LaTeX2e
\usepackage{iclr2026_conference,times}

\usepackage{amsmath,amsfonts,bm}

\def\eqref#1{equation~\ref{#1}}
\def\1{\bm{1}}

\DeclareMathAlphabet{\mathsfit}{\encodingdefault}{\sfdefault}{m}{sl}
\SetMathAlphabet{\mathsfit}{bold}{\encodingdefault}{\sfdefault}{bx}{n}

\usepackage[utf8]{inputenc} % allow utf-8 input
\usepackage[T1]{fontenc}    % use 8-bit T1 fonts
\usepackage{hyperref}       % hyperlinks
\usepackage{url}            % simple URL typesetting
\usepackage{booktabs}       % professional-quality tables
\usepackage{amsfonts}       % blackboard math symbols
\usepackage{nicefrac}       % for \nicefrac command
\usepackage{nicefrac}       % compact symbols for 1/2, etc.
\usepackage{microtype}      % microtypography
\usepackage{xcolor}         % colors

\usepackage{amsmath} % For math environments
\usepackage{amssymb} % For math symbols like \mathbb{R}
\usepackage{graphicx} % If you need to include figures

\usepackage{mathtools} % For \coloneqq
\usepackage[utf8]{inputenc}
\usepackage[english]{babel}

\usepackage{multirow} 
\usepackage{makecell}

\usepackage{subcaption}
\usepackage{wrapfig}
\usepackage{float}
\usepackage{pgffor}
\usepackage{adjustbox}
\usepackage{array}
\usepackage{caption}
\usepackage{rotating}

\usepackage{tcolorbox}
\tcbset{colback=gray!5, colframe=gray!80, boxrule=0.4pt}

\usepackage{appendix}

\DeclareMathOperator{\kmeans}{KMeans} % Or KNN if that's intended

\title{AutoMiSeg: Automatic Medical Image Segmentation via Test-Time Adaptation of Foundation Models}

\author{Xingjian Li\textsuperscript{1}\thanks{Equal contribution} \quad
  Qifeng Wu\textsuperscript{1}\footnotemark[1] \quad
  Adithya S. Ubaradka\textsuperscript{3} \quad
  Yiran Ding\textsuperscript{2} \quad
  Colleen Que\textsuperscript{1} \quad
   \\[0.55em]
  \textbf{Runmin Jiang\textsuperscript{1}} \quad
  \textbf{Jianhua Xing\textsuperscript{4}} \quad
  \textbf{Tianyang Wang\textsuperscript{5}} \quad
  \textbf{Min Xu\textsuperscript{1}\thanks{Corresponding author}} \\ \\
  \textsuperscript{1}Carnegie Mellon University \quad
  \textsuperscript{2}Brown University \quad
  \textsuperscript{3}National Institute of Technology Karnataka \\
  \textsuperscript{4}University of Pittsburgh \quad
  \textsuperscript{5}University of Alabama at Birmingham
}

\iclrfinalcopy % Uncomment for camera-ready version, but NOT for submission.
\begin{document}

\maketitle

\begin{abstract}

  % The abstract paragraph should be indented \nicefrac{1}{2}~inch (3~picas) on
  % both the left- and right-hand margins. Use 10~point type, with a vertical
  % spacing (leading) of 11~points.  The word \textbf{Abstract} must be centered,
  % bold, and in point size 12. Two line spaces precede the abstract. The abstract
  % must be limited to one paragraph.

 Medical image segmentation is vital for clinical diagnosis, yet current deep learning methods often demand extensive expert effort, i.e., either through annotating large training datasets or providing prompts at inference time for each new case. This paper introduces a zero-shot and automatic segmentation pipeline that combines off-the-shelf vision-language and segmentation foundation models. Given a medical image and a task definition (e.g., "segment the optic disc in an eye fundus image"), our method uses a grounding model to generate an initial bounding box, followed by a visual prompt boosting module that enhance the prompts, which are then processed by a promptable segmentation model to produce the final mask. To address the challenges of domain gap and result verification, we introduce a test-time adaptation framework featuring a set of learnable adaptors that align the medical inputs with foundation model representations. Its hyperparameters are optimized via Bayesian Optimization, guided by a proxy validation model without requiring ground-truth labels. Our pipeline offers an annotation-efficient and scalable solution for zero-shot medical image segmentation across diverse tasks. Our pipeline is evaluated on seven diverse medical imaging datasets and shows promising results. By proper decomposition and test-time adaptation, our fully automatic pipeline not only substantially surpasses the previously best-performing method, yielding a 69\% relative improvement in accuracy (Dice Score from 42.53 to 71.81), but also performs competitively with weakly-prompted interactive foundation models.

\end{abstract}

\section{Introduction}
\label{intro}
%Medical image segmentation plays a critical role in diagnosis and treatment planning, yet traditional approaches are time-consuming and prone to human error. Recent models like MedSAM and One-Prompt have improved segmentation performance, but they still require a certain level of user expertise — specifically, the ability to provide accurate spatial prompts such as bounding boxes. This can be a barrier for users who are not familiar with the exact location or appearance of the target anatomical structures.
Medical image segmentation plays a critical role in diagnosis and treatment planning~\citep{patil2013medical,litjens2017survey}. Artificial Intelligence has emerged as a transformative force in this domain, significantly enhancing the efficiency and accuracy of clinical workflows~\citep{Hesamian19DeepReview}. In particular, deep learning-based segmentation models have outperformed traditional computer vision techniques by leveraging large datasets and effective deep feature learning~\citep{Ronneberger15Unet, Zhou18UnetPlusPlus, Oktay18AttentionUnet}. Despite these advancements, most state-of-the-art models rely heavily on supervised learning, which requires extensive and high-quality annotations provided by medical experts~\citep{Papandreou15Weakly}. Furthermore, supervised models have poor scalability as they are constrained to the pre-defined classes and image domains supported by the training data.

In recent years, the emergence of vision foundation models like the Segment Anything Model (SAM)~\citep{Kirillov23SAM} offers promising new avenues for more efficient image segmentation. Unlike conventional methods that require separate training on each individual dataset, SAM enables general segmentation at inference time using simple prompts such as points or bounding boxes, significantly reducing the need for extensive labeled data. More advanced models such as MedSAM~\citep{Ma23MedSAM} and SAM-Med2D~\citep{cheng2023sam} further improve medical image segmentation by fine-tuning SAM on clinical data. However, they may still fail on certain datasets due to the wide variability in medical image modalities, low-contrast features, and subtle anatomical boundaries. Moreover, inference-time prompting still requires expert supervision, which becomes non-negligible when processing large batches of data.

%In this work, we propose a novel method that integrates large language models (LLMs) into the segmentation pipeline, enabling a more intuitive and accessible interaction paradigm. Instead of relying on spatial annotations, users can simply describe the desired region using natural language. Our system interprets these descriptions and generates segmentations accordingly, reducing the reliance on visual annotation skills. Furthermore, the model can provide feedback and analysis of the segmentation output, allowing users to iteratively refine their understanding and improve results through dialogue, rather than pixel-level input.

\begin{figure}[!t]
\centering
\includegraphics[width=\linewidth]{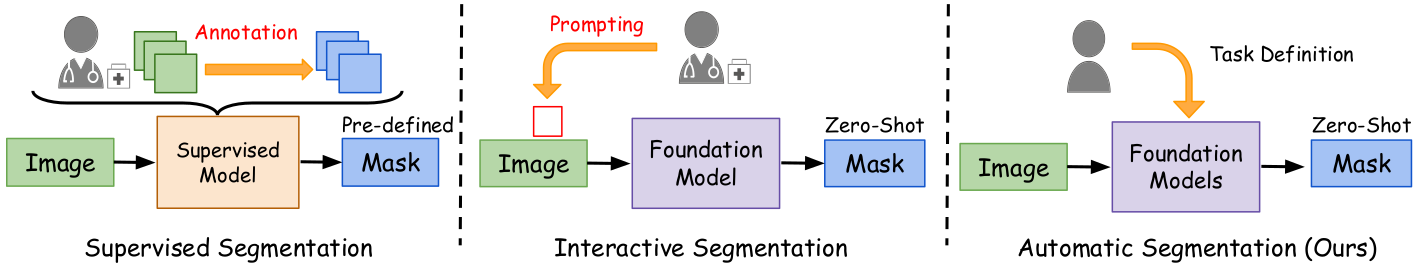}
\caption{Conceptually illustration of our proposed \textbf{annotation-free} and \textbf{non-interactive} paradigm for medical image segmentation (AutoMiSeg), compared with traditional supervised and interactive segmentation, which require considerable efforts of medical experts.}
\label{fig:concept}
\end{figure}

The goal of this paper is to realize a brand new paradigm of automatic medical image segmentation that embodies both the \emph{zero-shot} capacity of the prompting paradigm and the \emph{annotation-free inference} characteristic of the supervised learning paradigm. The idea is illustrated in Figure \ref{fig:concept}. To achieve this goal, the following key challenges must be addressed:

\noindent (1) Medical image segmentation is too complex for a single existing model to achieve effective zero-shot and annotation-free prediction. Therefore, properly decomposing the task into a sequence of simpler, modular steps is crucial for building a robust and adaptable pipeline.

\noindent (2) As medical images often exhibit domain-specific characteristics that differ from the data used to train general-purpose foundation models, effectively adapting medical images to the pre-trained models without losing their generalization capacity is essential for the broad applicability across various datasets. 

This paper presents a general pipeline \textbf{AutoMiSeg} for zero-shot, annotation-free automatic medical image segmentation. Given a medical image and a task description (e.g., "segment the optic disc from an eye fundus image"), the pipeline automatically generates the corresponding target mask without additional training or manual annotations. Our basic pipeline consists of a grounding module which aims to provide spatial prompts and an promptable segmentation module for the final mask generation. A prompt boosting module is integrated between them to enhance the prompt quality. Based on the pipeline, we propose a novel test-time adaptation method which employs Bayesian Optimization on a set of Learnable Test-time Adaptors (LTAs). A surrogate validation model is designed to evaluate the segmentation output, and its feedback is used to optimize the LTAs. 

Experiments across seven diverse medical imaging datasets demonstrate the effectiveness of AutoMiSeg. Our method delivers a 69\% relative improvement in accuracy, raising the average Dice Score from 42.53 up to 71.81, which represents a practically significant advancement over the strongest prior solution.
It also achieves competitive performance compared to weak-prompt foundation models, highlighting the potential of the new automatic segmentation paradigm. Detailed ablation studies confirm that the grounding model plays a critical role, and the LTAs significantly impact segmentation accuracy. Additionally, the proxy validator shows strong correlation with true segmentation performance, validating its use in guiding hyperparameter optimization.

%\textcolor{red}{To be updated according to experimental results.}
%In our experiments, we show that using bounding boxes and point prompts generated by a vision-language model (VLM), we can effectively guide powerful segmentation models such as SAM-HQ. This approach yields improved segmentation performance, as measured by Dice coefficients, compared to traditional prompting methods. In addition to quantitative improvements, our system provides natural language explanations alongside the segmentation results, offering interpretability and insight into the model's predictions.

\section{Related Work}
\label{rw}

\paragraph{Medical Image Segmentation.}
Deep learning, particularly U-Net architectures~\citep{Ronneberger15Unet, Zhou18UnetPlusPlus, Oktay18AttentionUnet}, has become the standard for medical image segmentation but is often hampered by the need for large, manually annotated datasets~\citep{Hesamian19DeepReview}. To reduce this dependency, researchers have explored few-shot~\citep{Snell17Prototypical}, weakly-supervised~\citep{Papandreou15Weakly}, and zero-shot learning (ZSL) methods~\citep{Xian19ZSLSurvey, xu2023side, yang2024sam}. Our work aligns with the emerging trend of \emph{training-free} methods that leverage large pre-trained foundation models directly for downstream tasks through prompting and composition, avoiding task-specific fine-tuning~\citep{Kirillov23SAM, Bommasani21FoundationModels}.

\paragraph{Promptable Models for Medical Image Segmentation.}
The Segment Anything Model (SAM)~\citep{Kirillov23SAM} initiated a paradigm shift towards promptable segmentation. While its zero-shot capabilities are impressive, studies in medical imaging found its performance inconsistent, often requiring precise user prompts and struggling with fine details or low-contrast regions~\citep{Mazurowski23SAMReview, He23SAMMedical, Deng23SAMReviewMed}. This led to medical-specific adaptations like MedSAM \citep{Ma23MedSAM}. Besides SAM based pipelines, One-Prompt~\citep{wu2024one} developed a customized network structure that accepts various types of prompt, such as doddle, box, click, etc, for medical image segmentation.  A key challenge remains the automation of prompt generation. Recent works have built text-driven pipelines, such as  SaLIP \citep{aleem2024test}, MedCLIP-SAM~\citep{koleilat2024medclip, koleilat2409medclip} and others aiming for universal text-prompted segmentation~\citep{zhao2023one}. In parallel, large-scale supervised models like BiomedParse~\citep{zhao2025foundation} achieve strong performance on a wide range of tasks but may lack the flexibility of zero-shot approaches for out-of-domain data. Our work is distinct in its fully automatic, training-free, and compositional design, which uses a dedicated grounding model to interpret text queries for initial localization, followed by a segmentation model.

\paragraph{Test-Time Adaptation.}
To handle domain shifts without retraining, Test-Time Adaptation (TTA) adapts models to new, unlabeled data during inference~\citep{Wang2020Tent, Sun2020TTT}. Recent approaches \citep{farina2024frustratingly,shin2024tta,hoang2024persistent} demonstrated effectiveness in adapting single backbones under distribution shift. However, most existing solutions rely on online backpropagation and hence are not suitable for non-differentiable pipelines. The most relevant work is SaLIP~\citep{aleem2024test}, which considers test-time adaptation but does not address the challenge of domain shift between medical images and general foundation models. Our work pioneers a novel TTA strategy specifically for a compositional segmentation pipeline. We introduce a set of Learnable Test-time Adaptors (LTAs) whose hyperparameters are optimized via Bayesian Optimization, guided by a proxy validation model, making our framework uniquely adaptable in a fully automatic, training-free manner.

\paragraph{AutoML for Medical Imaging.}
A key to the success of our method lies in applying AutoML principles \citep{he2021automl} to medical image segmentation. 
Existing work on AutoML for medical imaging \citep{jidney2023automl,ali2024review} has been applied to supervised models, focusing on network architecture design \citep{isensee2019nnu,yu2023eu} and hyperparameter configuration for training \citep{myronenko2023automated}. Despite boosting segmentation performance, these methods remain limited to task-specific supervised pipelines. To the best of our knowledge, no prior work has considered AutoML in the context of foundation model adaptation. Our work is the first to bridge this gap, offering a new direction towards robust and supervision-free medical imaging solutions.

\section{Methodology}
\label{met}
% \begin{enumerate}
%  \item Contrast improvement: in order to make the target region to stand out
%  \item Target area grounding: Prompt VLM with user input in text and image to be segmented to generate bounding boxes
%  \item Visual Prompt Augmentation:
%  \begin{enumerate}
%  \item Select the center of bbox as the anchor point
%  \item Use a pretrained vision encoder (DinoV2) in this case to encode an image
%  \item Choose top k (10 in my case) points that has highest similarity with the anchor point in feature space
%  \item Use KNN to cluster the 10 points into n groups (3 in my case)
%  \item Calculate the cluster center coordinate as the additional point prompts 
% \end{enumerate}
%  \item Promptable segmentation: boxes + points ==> segmentation mask
%  \item Feed the resulting mask and the original image into another ``Critic'' VLM to generate decent mask iteratively
% \end{enumerate}

\subsection{Overview}
\begin{figure}[!t]
\centering
\includegraphics[width=0.8\linewidth]{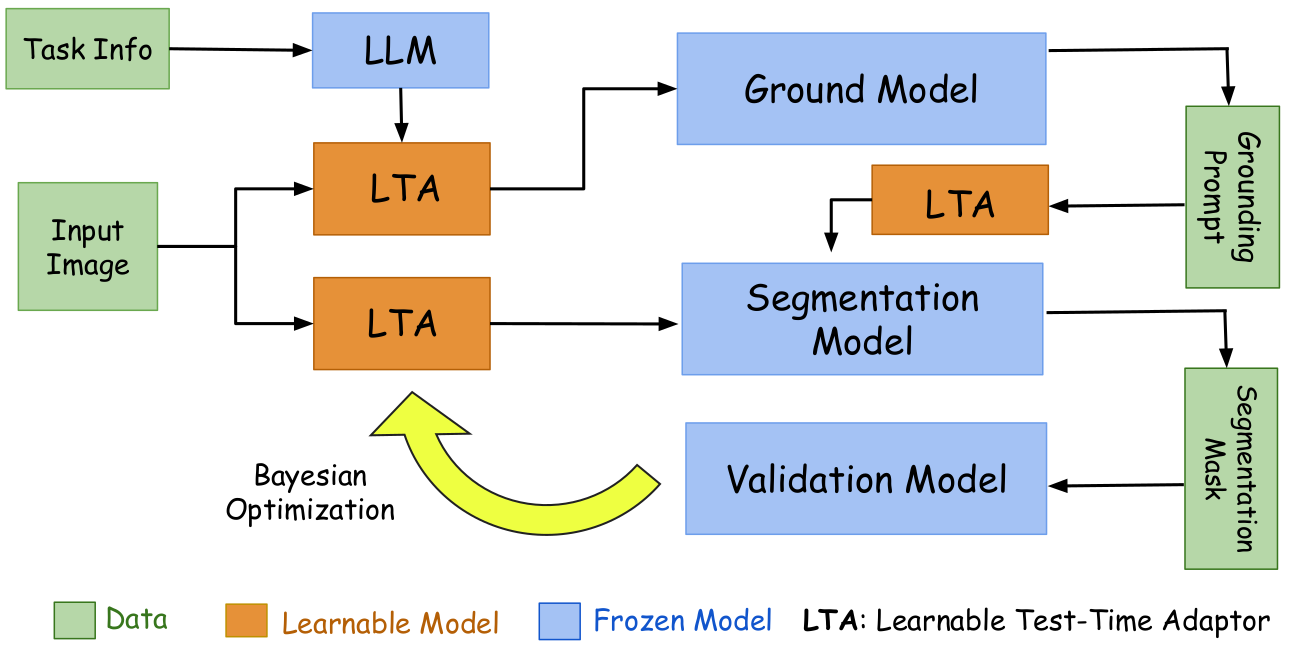}
\caption{Overview of the proposed AutoMiSeg pipeline. Given a medical image and task description, a grounding module predicts a coarse region, refined by a prompt booster. A promptable segmentation model then generates the binary mask. A validator checks consistency with the task, while Learnable Test-time Adaptors (LTAs) adapt inputs and are optimized via Bayesian Optimization.}
\label{fig:fig2}
%\vspace{-3mm}
\end{figure}

The AutoMiSeg pipeline achieves text-guided medical image segmentation without requiring any task-specific training or fine-tuning. The pipeline is illustrated in Figure \ref{fig:fig2}.

Let \( I \in \mathbb{R}^{H \times W \times C} \) denote a medical image and let \( T = \{T_{\text{target}}, T_{\text{whole}}\} \) denote a structured task description, where \( T_{\text{target}} \) specifies the object to be segmented (e.g., ``optic disc'') and \( T_{\text{whole}} \) names the image context (e.g., ``eye fundus image''). Given \( I \) and \( T \), AutoMiSeg outputs a binary segmentation mask \( M \in \mathbb{R}^{H \times W} \) that matches the task description. The pipeline begins with a \textbf{grounding module}, which takes the transformed image \( I_G \) and a sentence \(S_G\) constructed from the task definition $T$ as input and predicts a bounding box indicating the rough target region. The output is further refined by the \textbf{prompt booster}, which improves the prompt’s information and quality. The enhanced prompt is then passed to an \textbf{promptable segmentation module}, which also takes a transformed image \( I_S \) and generates the final segmentation mask \( M \).

As a fully automatic pipeline, we need a \textbf{validator} to estimate the quality of the output. It's realized by a vision-language model that checks whether the predicted mask \( M \) is consistent with the task description \( T \). To adapt the general pre-trained models to various medical image domains, we introduce \textbf{Learnable Test-time Adaptors (LTAs)}, which consists of a set of tunable operations applied to the inputs of the grounding and segmentation module. We apply \textbf{Bayesian Optimization} to search for the optimal configuration of LTAs that maximizes the validator's score. 

%It operates by sequentially composing several pre-trained foundation models, leveraging their inherent capabilities for localization, feature extraction, and segmentation. The pipeline processes an input medical image $I \in \mathbb{R}^{H \times W \times C}$ guided by a natural language text query $T$ describing the target region of interest. An overview is presented in Figure~\ref{fig:pipeline} [Reference Figure if you have one]. The core stages are detailed below.

\subsection{Target Area Grounding}
\label{subsec:grounding}
To initially localize the region specified by the text query $T = \{T_{\text{target}}, T_{\text{whole}}\}$, we leverage a pre-trained Vision-Language Model (VLM) capable of text-to-bounding-box grounding. Given the task definition, we first generate a set of prompt sentences $\{S_G|S_G=\mathbf{LLM} (T_{target}, T_{whole})\}$ which describe the visual characteristics of the target in the whole image. Here $\mathbf{LLM}$ refers to a general large language model which is responsible for generating a diverse set of text descriptions according to its internal medical knowledge about the image type and the target object. The input image $I$ is transformed to $I_G$ by a set of pre-defined vision operations with the specified hyperparameters. A grounding model $\mathrm{M_{grd}} (I_G, S_G)$ is then employed to generate a bounding box $B = (x_{min}, y_{min}, x_{max}, y_{max})$, which defines the coordinates of the predicted primary region corresponding to $T_{target}$. By default, we use ChatGPT-4o \citep{chatgpt} as the $\mathbf{LLM}$ and CogVLM \citep{wang2024cogvlm} as the grounding module. A more detailed introduction of grounding models and CogVLM is presented in Appendix A.

\subsection{Visual Prompt Boosting}
\label{subsec:prompt_augmentation}
Inspired by CoVP \citep{tang2024chain}, to refine the initial bounding box $B$, we generate $n$ supplementary positive point prompts via a deterministic, feature-driven process. This utilizes a pre-trained DINOv2 \citep{Oquab24DINOv2}, a pretrained vision foundation model valued for its strong semantic features representations.
First, an anchor point $p_a$ is set to the geometric center of $B$, and its DINOv2 feature vector $f_a$ is extracted. We then identify the top-$k$ points $P_k$ (e.g., $k=10$), restricted to lie within $B$ for efficiency, whose DINOv2 features $f(p)$ exhibit the highest cosine similarity to $f_a$.
To ensure these $P_k$ points represent diverse parts of the target object, their spatial coordinates are subsequently clustered into $n$ groups (e.g., $n=3$) using the K-Means algorithm. The centroids of these $n$ clusters constitute the final set of generated positive point prompts, $P_c$. These points $P_c$ are then used in conjunction with the original bounding box $B$ by the promptable segmentation model. A more detailed description of the visual prompt boosting module is presented in Appendix B. 

\subsection{Promptable Segmentation}
\label{subsec:segmentation}
Equipped with the bounding box $B$ from Section~\ref{subsec:grounding} and the augmented point set $P_c$ generated in Section~\ref{subsec:prompt_augmentation}, we utilize a pre-trained promptable segmentation model $\mathrm{M}_{seg}$, such as the Segment Anything Model (SAM) \citep{Kirillov23SAM}. The model takes the image $I$ and the visual prompts $(B, P_c)$ as input and produces a segmentation mask $M$ corresponding to the prompted object, again without any training or fine-tuning.
\begin{equation}
    M = \mathrm{M}_{seg}(I, B, P_c)
    \label{eq:sam_segmentation}
\end{equation}

%\subsection{Parameterized Contrast Improvement (Optional Pre-processing)}
%\label{subsec:contrast}
%Medical images, particularly from modalities like CT or MRI, can exhibit varying contrast levels, potentially obscuring the target region. As an optional initial step, we apply a standard parameterized contrast enhancement technique (e.g., Contrast Limited Adaptive Histogram Equalization - CLAHE \citep{Zuiderveld94CLAHE} or simple intensity windowing based on domain knowledge) to $I$. This aims to improve the visibility and distinctiveness of the target structure specified in $T$, potentially aiding subsequent localization and feature extraction steps. Let the enhanced image be denoted as $I'$. For simplicity, we refer to the input image as $I$ in subsequent steps, assuming enhancement has been applied if deemed necessary.

\subsection{Learnable Test-time Adaptors}
We consider the following two types of tunable variables in our test-time adaptors. A complete description about the variable types and value ranges are introduced in Appendix C. 

\paragraph{Domain-adapted Image Transformation.} To enhance the compatibility of medical images with general pre-trained vision models, we applied a set of targeted image transformations aimed at improving visual quality and mitigating domain-specific biases. Each transformation was independently applied to both the grounding and segmentation models using separate hyperparameters to optimize performance for their respective tasks. The operations specifically include (1) \textbf{HSV shift}, which is used to simulate variations in lighting and acquisition conditions; (2) \textbf{Channel-wise RGB shift}, which introduces color perturbations that reduce reliance on fixed intensity distributions often found in medical datasets; (3) \textbf{Contrast Limited Adaptive Histogram Equalization (CLAHE)} \citep{zuiderveld1994contrast}, which is applied to enhance local contrast and reveal subtle anatomical structures typically underrepresented in low-contrast regions; and (4) \textbf{unsharp masking}, which is employed to emphasize fine details and edge clarity, which are critical for precise localization and segmentation. These transformations collectively promote better alignment between the test images and the pre-trained vision models.

\paragraph{Automatic Prompt Enhancement.}LTAs also incorporate an automatic prompt enhancement mechanism to better guide the grounding and segmentation models. Specifically, we consider two tunable choices: (1) selecting the most effective prompt sentence for the grounding model from a set of candidates generated by the LLM, and (2) determining the optimal value of $k$ in the prompt boosting module, which controls how many points with the most similar features are aggregated to produce the final spatial prompt. 
 
\subsection{Proxy Validator}
The proxy validator is designed to estimate the chance that a segmentation mask generated by $\mathbf{M}_{seg}$ is the actual target. While it is not a rigorous probability score, we expect that a higher validation score indicates better quality of the generated mask. The designed proxy validator specifically consists of the following two pseudo evaluation tasks. 

\paragraph{Pseudo Evaluation with Zero-shot Classification.}Given an image $I$ and the candidate segmentation mask $M$, we first generate the test image by only keeping the target region as $I_{test} = I \odot (1-M)$, where $\odot$ denotes pixel-wise multiplication. To design the pseudo classification task, we use a predefined template incorporating the task information $T_{target}$ and $T_{whole}$ to prompt a general LLM, and get a list of contrastive categories in the form of text descriptions $\{T_{c_i}\}_{i=1}^m$. We then employ a vision-language model $\mathrm{M}_{val}$ to perform zero-shot classification for $I_{test}$ and output the probability score for $T_{target}$ as the validation score $S_{zc}$.

\paragraph{Pseudo Evaluation with Image-text Matching}While the zero-shot classification evaluates the image-text alignment with formal medical terminologies, it doesn't consider to verify the segmented region by its expected visual characteristics such as color, shape, textures and so on. Here we employ another pseudo evaluation by considering the image-text matching. Similar to the grounding module, we use a LLM to generate a set of descriptive prompts, and apply the same $\mathrm{M}_{val}$ to calculate the similarity between the test image $I_{test}$ and each descriptor. The average image-text matching score $S_{mt}$ is calculated and contributed to the validation score. The final validation score for an input image $I$ and its candidate segmentation $M$ is $S_{val} = S_{zc}+S_{mt}$. By default, we employ BioMedCLIP \citep{zhang2024biomedclip} as $\mathrm{M}_{val}$. It is worth noting that, with the pre-defined LLM templates, the validation process is fully automatic given the task definition $T=(T_{target}, T_{whole})$ as input. In Appendix D, we present all the LLM templates used in our pipeline.

\subsection{Test-Time Adaptation}
To make the domain-adapted pipeline compatible to black box modules (e.g., foundation models with API interfaces) and alleviate potential over-fitting, we adopt the Bayesian Optimization \citep{snoek2012practical} in our pipeline for test-time adaptation. The goal is to maximize the evaluation score of the proxy validator by tuning the hyperparameters used in the LTA modules. Specifically, we employ the Tree-structured Parzen Estimator (TPE) as the surrogate model in Bayesian Optimization to better support mixed-type variables. To balance the efficiency and effectiveness, we allow up to $N_t$ trials (e.g., $N_t=100$) to identify the optimal configuration on a subset of the test set up to $N_s$ (e.g., $N_s=100$) examples for Bayesian Optimization. The solved configuration is then applied to the entire test set.  
\begin{table}[t]
\caption{Segmentation performance comparison across seven medical imaging datasets. 
Reported numbers are Dice scores. MedSAM results on weak prompts are ignored as they are extremely low.}
\label{tab:performance_on_datasets}
\centering
\small
\renewcommand{\arraystretch}{1.1}
\adjustbox{max width=\textwidth}{
\begin{tabular}{c|c|ccccccc|c}
\toprule
\multirow{2}{*}{\textbf{Category}} & \multirow{2}{*}{\textbf{Seg model}} & 
\multirow{2}{*}{\textbf{Kvasir }} & 
\multirow{2}{*}{\textbf{Busi}} & 
\multirow{2}{*}{\textbf{Isic2016 }} & 
\multirow{2}{*}{\textbf{Promise12 }} & 
\multirow{2}{*}{\textbf{Kidney}} & 
\multirow{2}{*}{\textbf{SkinCancer}} & 
\multirow{2}{*}{\textbf{REFUGE}} & 
\multirow{2}{*}{\textbf{Average}} \\
& & & & & & & & & \\
\midrule
\multirow{6}{*}{\textbf{Supervised}} 
  & ResNet-18 \citep{resnet}  & 73.9 & 67.4 & 87.8 & 89.5 & 97.9 & 86.4 & 90.4 & 84.8 \\
  & ResNet-50 \citep{resnet}  & 69.8 & 63.2 & 88.7 & 88.8 & 97.8 & 84.6 & 90.1 & 83.3 \\
  & EfficientNet \citep{efficientnet} & 81.2 & 71.1 & 90.3 & 89.2 & 98.1 & 89.0 & 88.2 & 86.7 \\
  & MobileNet-v2 \citep{mobilenetv2} & 75.4 & 65.5 & 89.1 & 89.6 & 98.0 & 87.9 & 84.5 & 84.3 \\
  & DenseNet-121 \citep{densenet} & 79.4 & 69.5 & 89.3 & 90.0 & 98.0 & 85.6 & 91.1 & 86.1 \\
  & Mix-ViT \citep{mixvit} & 56.9 & -- & 89.1 & -- & -- & 88.1 & 88.1 & -- \\
  & BiomedParse \citep{zhao2025foundation} & 90.7 & 87.4 & 94.3 & 84.5 & 78.9 & 95.6 & 78.5 & 87.1 \\
\midrule
\multirow{3}{*}{\textbf{Strong prompt}} 
  & SAM-Med2D \citep{cheng2023sam} & 89.63 & 89.91 & 93.89 & 87.00 & 87.82 & 92.99 & 83.49 & 89.25 \\
  & MedSAM \citep{Ma23MedSAM} & 96.46 & 92.40 & 92.88 & 88.77 & 97.02 & 95.79 & 91.35 & 93.52 \\
  & SAM \citep{Kirillov23SAM} & 94.83 & 87.41 & 87.38 & 91.10 & 91.97 & 93.59 & 91.89 & 91.17 \\
\midrule
\multirow{2}{*}{\textbf{Weak prompt}} 
  %& SAM-Med2D \citep{cheng2023sam} & 61.92{\scriptsize±1.37} & 76.39{\scriptsize±0.94} & 87.22{\scriptsize±0.19} & 62.11{\scriptsize±0.75} & 56.81{\scriptsize±0.55} & 88.18{\scriptsize±0.93} & 52.74{\scriptsize±0.61} & 69.33{\scriptsize±0.76} \\
  & SAM-Med2D \citep{cheng2023sam} & 61.92 & 76.39 & 87.22 & 62.11 & 56.81 & 88.18 & 52.74 & 69.33 \\
  %& SAM \citep{Kirillov23SAM} & 86.37{\scriptsize±0.86} & 65.39{\scriptsize±1.72} & 72.24{\scriptsize±1.45} & 65.09{\scriptsize±0.35} & 70.91{\scriptsize±0.62} & 88.33{\scriptsize±1.17} & 74.32{\scriptsize±0.69} & 74.66{\scriptsize±0.98} \\
  & SAM \citep{Kirillov23SAM} & 86.37 & 65.39 & 72.24 & 65.09& 70.91 & 88.33 & 74.32& 74.66 \\
\midrule
\multirow{4}{*}{\textbf{Automatic zero-shot}}
%  & SaLIP & 32.35{\scriptsize±9.74} & 19.39{\scriptsize±1.44} & 28.05{\scriptsize±4.73} & 19.21{\scriptsize±2.25} & 8.15{\scriptsize±2.56} & 48.66{\scriptsize±9.27} & 29.06{\scriptsize±3.81} & 26.41{\scriptsize±4.82} \\
  & SaLIP \citep{aleem2024test} & 32.35 & 19.39 & 28.05& 19.21 & 8.15& 48.66 & 29.06 & 26.41 \\
  & MedCLIP-SAM \citep{koleilat2024medclip} & 56.29 & 10.60 & 36.53 & 2.49 & 6.49 & 48.23 & 4.58 & 23.26 \\
  & MedCLIP-SAM-v2 \citep{koleilat2409medclip} & 44.57 & 34.26 & 47.33 & 17.15 & 32.55 & 74.52 & 47.35 & 42.53 \\
  & AutoMiSeg \textbf{(Ours)} & \textbf{74.80} & \textbf{61.65} & \textbf{68.38} & \textbf{60.61} & \textbf{73.05} & \textbf{84.41} & \textbf{79.78} & \textbf{71.81} \\
\bottomrule
\end{tabular}
}
\end{table}

\section{Experiments}
\label{exp}
\subsection{Datasets}
To ensure a comprehensive and diverse evaluation of segmentation performance, we collected seven public medical imaging datasets. These datasets span a broad range of imaging modalities, including fundus photography, endoscopy, ultrasound, dermoscopy and MRI. They cover multiple organ systems such as the eyes, gastrointestinal tract, breast, skin, prostate and kidneys, enabling an assessment of generalization across heterogeneous domains. %The selected datasets encompass both anatomical and pathological targets, ranging from polyp and tumor segmentation to anatomical boundary detection, providing a rigorous testbed for evaluating zero-shot segmentation capabilities.

Here we provide a brief introduction of the datasets. (1) The REFUGE \citep{orlando2020refuge} dataset includes retinal fundus photographs used to segment the optic disc for glaucoma risk assessment. (2) The Kvasir \citep{Pogorelov:2017:KMI:3083187.3083212}dataset contains endoscopic images of the gastrointestinal tract with annotated polyps. (3) The Busi \citep{walid_al-dhabyani_mohammed_gomaa_hussien_khaled_aly_fahmy_2024} dataset provides ultrasound images of the breast labeled for benign and malignant tumors. (4) The ISIC2016 \citep{gutman2016skin} dataset includes dermoscopic images annotated for general skin lesion segmentation. (5) The UWSkinCancer \citep{uw_skin_cancer_2021} dataset focuses on skin cancer detection from dermoscopic images. (6) The Promise12 \citep{litjens2014evaluation} dataset comprises T2-weighted prostate MRI scans annotated for prostate volume segmentation. (7) The Usforkidney \citep{song2022ct2us} \citep{ct2usforkidneyseg2024} dataset offers kidney ultrasound images labeled for tumor segmentation. Except for REFUGE \citep{orlando2020refuge}, all the remaining datasets are from MedSegBench \citep{Ku2024} and we adopt the official test splits. 

\subsection{Evaluation}
We evaluate our AutoMiSeg pipeline on the official test split of each dataset to ensure a fair and consistent comparison. As our method represents the first fully automatic, non-interactive zero-shot segmentation framework in the medical imaging domain, we additionally benchmark it against both supervised models trained on pre-defined classes and interactive foundation models guided by various types of prompts. These competing methods serve as practical \emph{upper bounds} for our pipeline, as they rely on different forms of expert intervention, such as labeled data or inference-time guidance, which our approach explicitly avoids.

For supervised models, we present the performance reported in the MedSegBench \citep{Ku2024} paper. For interactive models, we follow the practice of SAM-Med2D \citep{cheng2023sam} and One-prompt \citep{wu2024one} to simulate the interaction process. Specifically, we consider two types of interaction, which are strong and weak prompts. The strong prompt is derived from the ground truth mask as the minimal bounding box enclosing all foreground pixels, defined by its top-left and bottom-right coordinates. The weak prompt is defined as a single point uniformly sampled from the target region's foreground pixels. We also evaluate baseline methods SaLIP \citep{aleem2024test}, MedCLIP-SAM \citep{koleilat2024medclip} and MedCLIP-SAM-v2 \citep{koleilat2409medclip}, which are also designed to achieve zero-shot medical image segmentation. 

\begin{figure}[t]
%\vspace{-5mm}
    \centering
    \includegraphics[width=0.98\textwidth]{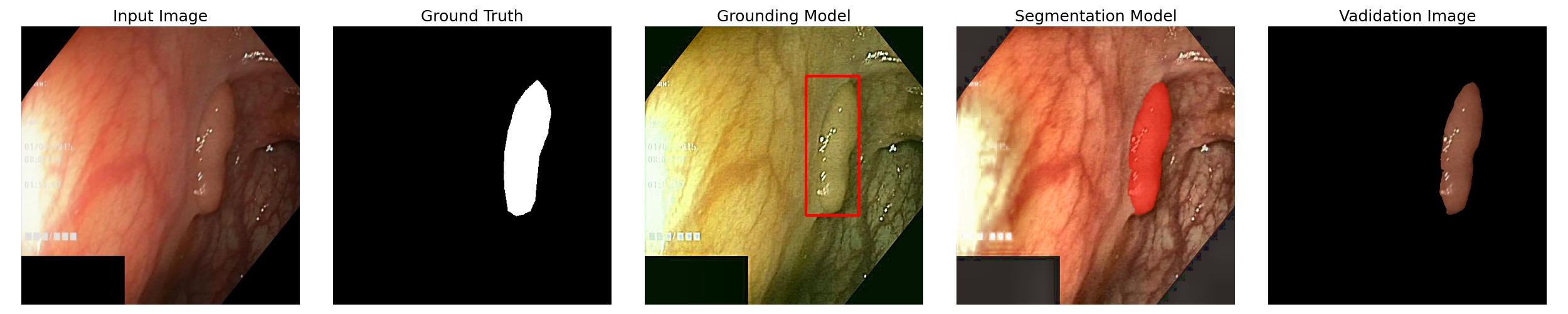}
 
    \includegraphics[width=0.98\textwidth]{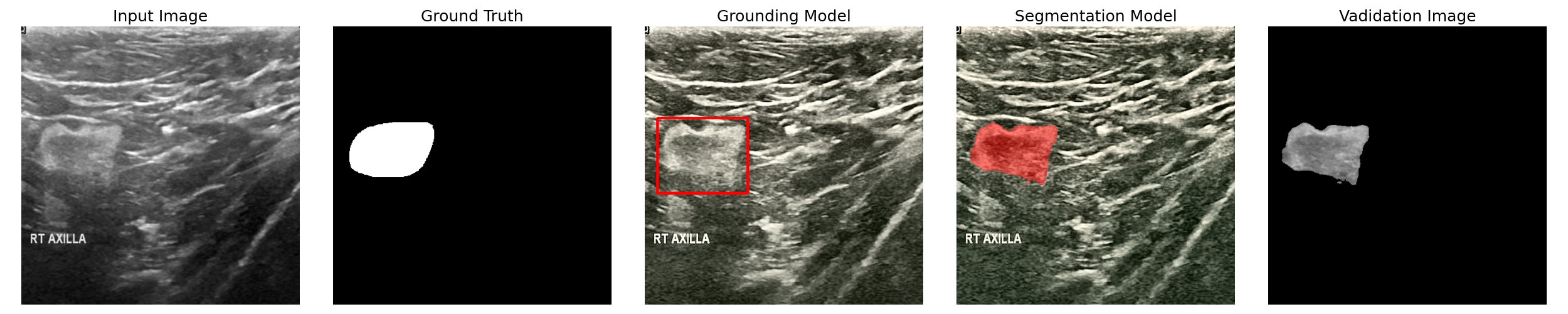}
    
    \caption{Qualitative visualization of our AutoMiSeg pipeline on two examples from the Kvasir \citep{Pogorelov:2017:KMI:3083187.3083212} and Busi \citep{walid_al-dhabyani_mohammed_gomaa_hussien_khaled_aly_fahmy_2024} dataset. The columns from left to right represent: (1) input image, (2) ground truth, (3) input image of the grounding model and its output, (4) input of the segmentation model and its output, and (5) input of the validator. }
    \label{fig:demo_cases}
\end{figure}

\subsection{Main Results}
From the results shown in Table \ref{tab:performance_on_datasets}, supervised models achieve consistently strong results and interactive foundation models with strong prompts demonstrate even higher or comparable performance. %This highlights the impressive generalization ability of foundation models when provided with high-quality guidance. 
However, models using weak prompts exhibit a notable drop in performance. This emphasizes the importance of our proposed strategy that decomposes the segmentation task and incorporates a grounding model to first localize the target region with a bounding box. Within the zero-shot setting group, our method AutoMiSeg substantially surpasses the previously best-performing method MedCLIP-SAM-v2 \citep{koleilat2409medclip} by 69\% relative improvement, i.e., improving the average dice score from 42.53 to 71.81. Figure \ref{fig:demo_cases} shows qualitative examples of our results. More examples are presented in Appendix E. 

Interestingly, within the weak prompt group, SAM \citep{Kirillov23SAM} and SAM-Med2D \citep{cheng2023sam} do not exhibit consistent relative performance across all datasets. For instance, SAM \citep{Kirillov23SAM} significantly outperforms SAM-Med2D \citep{cheng2023sam} on Kvasir \citep{Pogorelov:2017:KMI:3083187.3083212} (87.21 vs. 61.49), whereas SAM-Med2D \citep{cheng2023sam} yields higher accuracy on Busi \citep{walid_al-dhabyani_mohammed_gomaa_hussien_khaled_aly_fahmy_2024} (77.99 vs. 68.50). These discrepancies suggest that different models (even particularly adapted with medical data such as SAM-Med2D \citep{cheng2023sam}) have dataset-specific strengths and are affected differently by prompt ambiguity, reinforcing the need for test-time adaptation in zero-shot settings.

Our fully automatic pipeline achieves comparable performance with interactive foundation models under weak prompts. While it does not yet reach the performance levels of supervised models or strong-prompt foundation models, this result underscores the promise of the automatic, zero-shot segmentation paradigm\footnote{Note that BiomedParse~\citep{zhao2025foundation} also supports text input for segmentation. Although this capability resembles a zero-shot paradigm, these approaches still fundamentally fall under the supervised category. A more detailed discussion is provided in Appendix F.}. By eliminating the need for manual prompts and annotations, our approach provides a scalable solution that balances performance with usability, and demonstrates the potential of further research in this direction. Regarding the efficiency, our method shows comparable inference speed with existing zero-shot pipelines. Detailed results are in Appendix G.

\section{Mechanism Analysis}
\label{ana}

\begin{table}[t]
\caption{Performance comparison across different combinations of $P_{\text{grd}}$ and $P_{\text{seg}}$ on Kvasir \citep{Pogorelov:2017:KMI:3083187.3083212} and Busi \citep{walid_al-dhabyani_mohammed_gomaa_hussien_khaled_aly_fahmy_2024} datasets. 
In these experiments, we fix the optimal choice of the prompt enhancement hyperparameters and only affect the domain-adapted image transformation process for simplicity. 
Base refers to using the original image without any transformation.}
\label{tab:performance_comparison}
\centering
\small
\renewcommand{\arraystretch}{1.1}
\adjustbox{max width=\textwidth}{
\begin{tabular}{llcc}
\toprule
$P_{\text{grd}}$ & $P_{\text{seg}}$ & Kvasir \citep{Pogorelov:2017:KMI:3083187.3083212} (endoscopy) & 
Busi \citep{walid_al-dhabyani_mohammed_gomaa_hussien_khaled_aly_fahmy_2024} (ultrasound) \\
\midrule
Optimal & Optimal & 74.80 & 61.65 \\
Optimal & Base    & 71.43 & 57.13 \\
Optimal & Random  & 66.97 & 54.86 \\
Base    & Optimal & 25.94 & 15.72 \\
Random  & Optimal & 22.78 & 15.09 \\
\bottomrule
\end{tabular}
}
\end{table}

\subsection{Role of the Grounding Model}
We find that the grounding model is the most critical component in the pipeline. This is expected, since the segmentation model relies on the grounding box to localize the target. %If the box is inaccurate, the segmentation output will be unreliable regardless of the segmentation model's quality. 
The results show a dramatic drop in performance when the grounding model uses base or random transformations. For example, on the Kvasir dataset, the Dice score drops from 74.80 (Optimal) to 25.94 (Base) and 22.78 (Random). This confirms that test-time image transformation is essential for grounding accuracy and the reliability of the whole pipeline. When the grounding model is optimal, the segmentation model is more robust to changes. Even with random or no transformation, the performance only drops slightly. This also aligns with the observations in Table \ref{tab:performance_comparison} that foundation models with strong box prompts achieve excellent performance.  
Interestingly, random transformations perform worse than using the original image. This suggests that poorly chosen transformations may hurt performance more than help. Careful searching the hyperparameter space is necessary to achieve good results.

\begin{figure}[t]
    \centering
    \begin{subfigure}[t]{0.24\textwidth}
        \includegraphics[width=\linewidth]{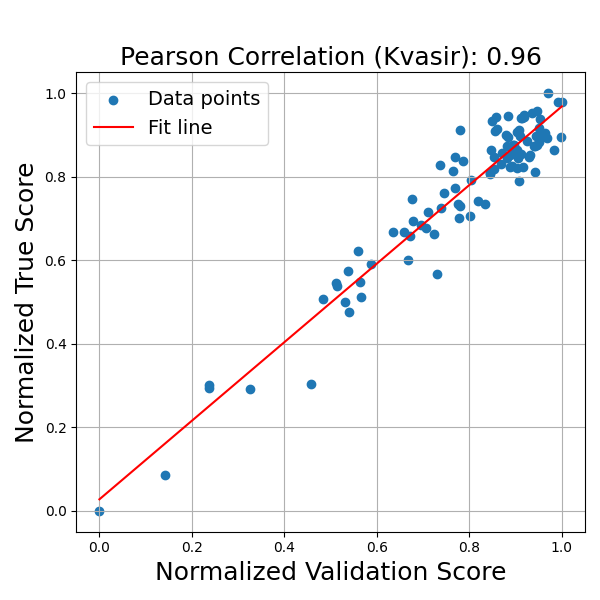}
    \end{subfigure}
    \hfill
    \begin{subfigure}[t]{0.24\textwidth}
        \includegraphics[width=\linewidth]{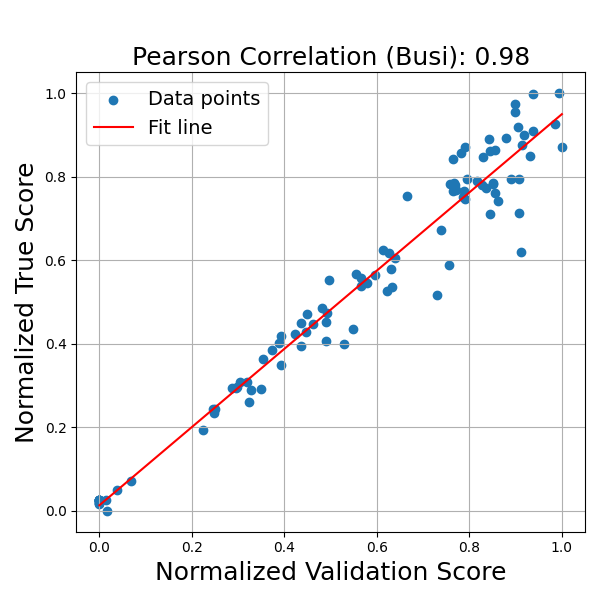}
    \end{subfigure}
    \hfill
    \begin{subfigure}[t]{0.24\textwidth}
        \includegraphics[width=\linewidth]{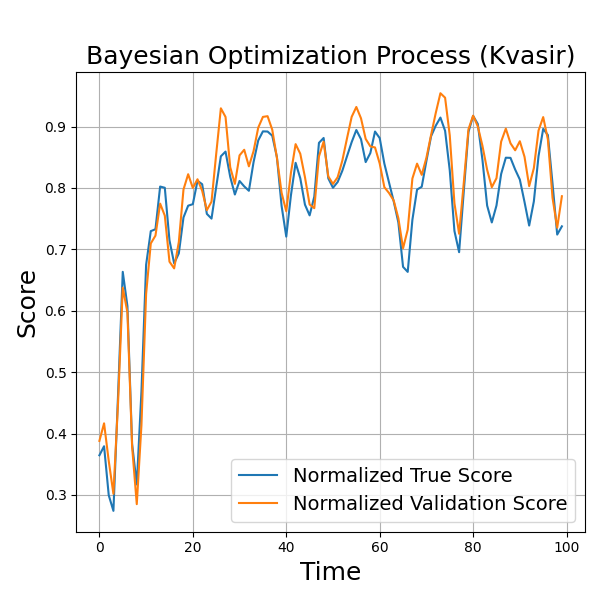}
    \end{subfigure}
    \hfill
    \begin{subfigure}[t]{0.24\textwidth}
        \includegraphics[width=\linewidth]{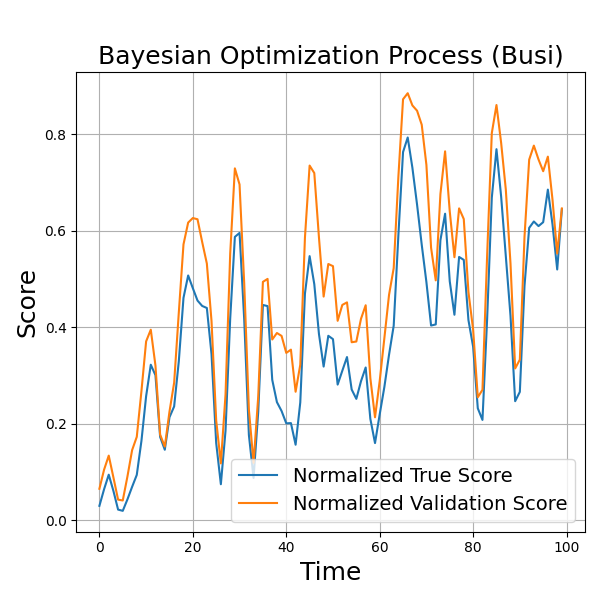}
    \end{subfigure}
    \caption{The left two figures show correlation between the normalized validation scores and true scores on the Kvasir \citep{Pogorelov:2017:KMI:3083187.3083212} Busi \citep{walid_al-dhabyani_mohammed_gomaa_hussien_khaled_aly_fahmy_2024} datasets with the searched hyperparameters. The right two figures present the process of Bayesian Optimization.}
    \label{fig:validator_perf}
%\vspace{-3mm}
\end{figure}

\begin{figure}[t]
\centering
\includegraphics[width=\linewidth]{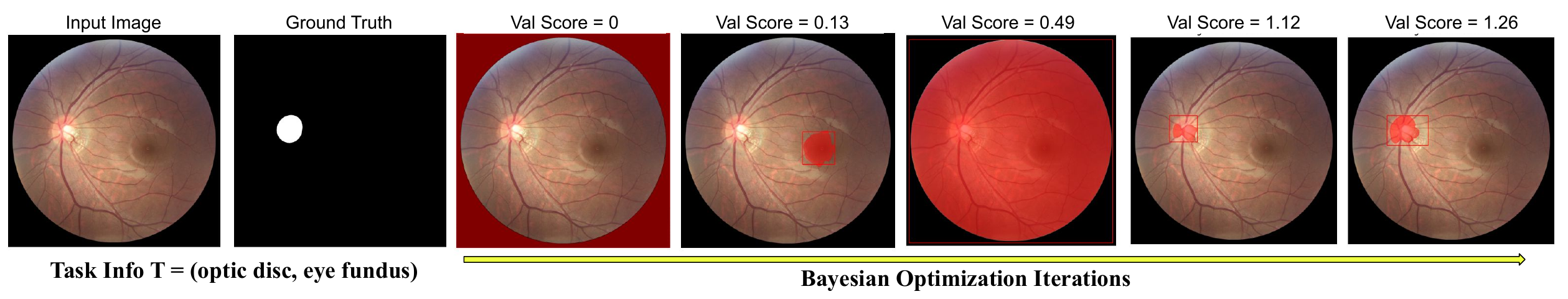}
\caption{An example from the REFUGE \citep{orlando2020refuge} dataset showing how the segmentation quality gradually improves in the optimization process. }
\label{fig:showcase-refuge}
%\vspace{-3mm}
\end{figure}

\subsection{Performance of the Proxy Validator}

Additional analyses are presented to verify both the quality of the proxy validator and the dynamics of the Bayesian optimization process. In the analyses, ground truth masks are used post hoc to assess the alignment between the validation scores and the actual performance (measured by Dice score), without influencing the optimization directly. 
As we can observe in Figure \ref{fig:validator_perf}, the two left plots reveal a strong linear relationship between the normalized validation scores and true scores on the Kvasir \citep{Pogorelov:2017:KMI:3083187.3083212} and Busi \citep{walid_al-dhabyani_mohammed_gomaa_hussien_khaled_aly_fahmy_2024} datasets, with high Pearson correlation coefficients, confirming that the designed validator serves as a reliable surrogate for model performance. Meanwhile, the two right plots show how the validation and true scores evolves during Bayesian Optimization. The overall trends clearly move toward higher-performing configurations, validating the effectiveness of the search process. Some fluctuations (especially for the Busi \citep{walid_al-dhabyani_mohammed_gomaa_hussien_khaled_aly_fahmy_2024} dataset) reflect the inherent trade-off between exploring new hyperparameter regions and exploiting promising ones.

Figure \ref{fig:showcase-refuge} illustrates a representative case from the REFUGE \citep{orlando2020refuge} dataset, where the task is segmenting the optic disc from a fundus image. The optimization process begins with random configurations in the LTA space, often resulting in the segmentation of background regions. In early stages, the macula, which has a similar shape and size to the optic disc, is mistakenly segmented. However, the validator assigns low zero-shot classification and image-text matching scores to these incorrect predictions. As the optimization progresses, the pipeline gradually refines its focus, ultimately localizing the optic disc and producing accurate segmentation results.

\begin{wrapfigure}{r}{0.35\linewidth}
%\vspace{-8mm}
    \centering
    \includegraphics[width=0.98\linewidth]{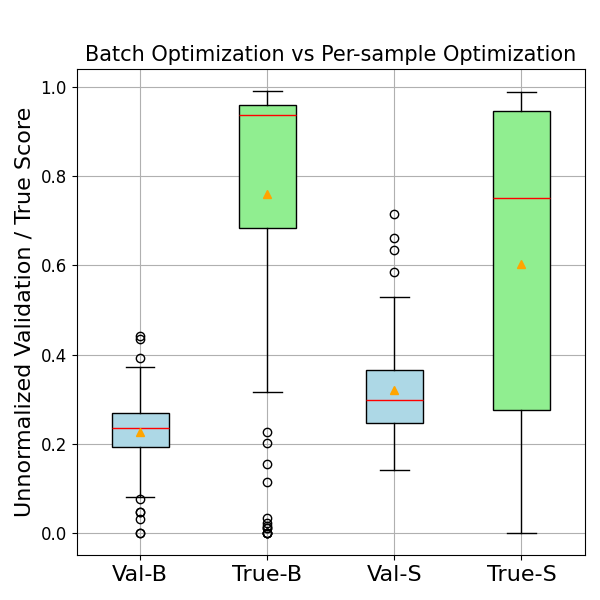}
    \caption{Evaluate per-sample BO on the Kvasir dataset. B refers to batch and S refers to per-sample.}
    \label{fig:per-sample}
\vspace{-8mm}
\end{wrapfigure}

We also find hyperparameter selection with BO on a batch of test samples show better performance than per-sample BO. As shown in Figure \ref{fig:per-sample}, although per-sample BO achieves higher validation scores, the overall true scores decrease compared to the default setting of batch BO, as BO tends to overfit each individual test example and may lose the chance to find more robust configurations. 

\section{Limitation and Future Work}
While AutoMiSeg substantially outperforms prior automatic zero-shot segmentation methods, a clear performance gap remains compared to supervised models and foundation models guided by strong expert prompts. Given the modular nature of our pipeline, a promising future direction is to expand the search space to include different model configurations. As shown in our preliminary ablation studies (Appendix H), there is considerable room for improvement through better foundation model selection within the AutoML framework. Our method will also naturally benefit from the continuous advances of general foundation models.
Another important avenue is to develop more effective test-time adaptation strategies tailored for the non-differentiable and compositional nature of the pipeline.

%\section{Limitation and Future Work}
%Despite demonstrating the potential of a fully automatic, zero-shot segmentation framework, our method has several limitations that warrant further investigation.

%First, the proposed pipeline heavily relies on the performance of the grounding model to generate accurate spatial prompts. As the grounding module serves as the initial step in localizing the target region, any inaccuracies significantly affect the downstream segmentation quality. While we adopt a state-of-the-art grounding model in our framework, more powerful and robust grounding models  (particularly those designed for medical domains) will be employed in future to enhance the overall pipeline performance. 

%Second, our current framework is designed specifically for binary segmentation tasks, where each instance involves isolating a single target structure from the background. Although multi-class or multi-instance segmentation could be approached by sequentially prompting each target, similar to interactive methods, this strategy introduces additional challenges in the context of a fully automatic system. In particular, potential conflicts may arise during the generation of overlapping box prompts or in the final fusion of multiple instance masks. Addressing these issues will require new strategies for prompt disambiguation and post-processing to ensure semantic consistency and spatial coherence across multiple targets.

\section{Conclusion}
\label{con}
This paper presents AutoMiSeg, a fully automatic, zero-shot segmentation pipeline for medical images that eliminates the need for manual annotations or interaction at inference time. By leveraging pretrained vision-language and segmentation foundation models, along with a novel test-time adaptation framework, our method effectively addresses the challenges of domain shift and prompt generation in medical image segmentation. Experimental results across seven diverse datasets demonstrate the pipeline's significant advantages over zero-shot baselines and also competitive performance compared to both weak-prompt interactive models. 
This work underscores the potential of modular, training-free pipelines for scalable medical image analysis and opens up promising directions for future research in improving the performance with more powerful pre-trained models and designing advanced test-time adaptation approaches for more complex pipelines. 

%In this work, we present a novel approach to medical image segmentation that reduces the reliance on expert-driven spatial prompts by integrating vision-language models (VLMs) into the segmentation pipeline. By leveraging natural language input to generate bounding boxes and point prompts, our method enables users to interact with the system using semantic descriptions rather than manual annotations. We demonstrate that the combination of VLM-generated prompts and powerful segmentation models like SAM-HQ leads to improved performance, as measured by Dice coefficients, across multiple datasets. Furthermore, our framework provides interpretability through natural language feedback, facilitating a more accessible and explainable segmentation process. This work highlights the potential of combining VLMs and segmentation models to democratize medical image analysis and open new directions for human-AI collaboration in clinical workflows.

\section*{Acknowledgement}
This study was partially funded by U.S. NSF grants MCB-2205148.

\bibliography{iclr2026_conference}

\begin{thebibliography}{61}
\providecommand{\natexlab}[1]{#1}
\providecommand{\url}[1]{\texttt{#1}}
\expandafter\ifx\csname urlstyle\endcsname\relax
  \providecommand{\doi}[1]{doi: #1}\else
  \providecommand{\doi}{doi: \begingroup \urlstyle{rm}\Url}\fi

\bibitem[Al-Dhabyani et~al.(2024)Al-Dhabyani, Gomaa, Khaled, and Fahmy]{walid_al-dhabyani_mohammed_gomaa_hussien_khaled_aly_fahmy_2024}
Walid Al-Dhabyani, Mohammed Gomaa, Hussien Khaled, and Aly Fahmy.
\newblock Busi (breast ultrasound images dataset), 2024.
\newblock URL \url{https://www.kaggle.com/ds/4519784}.

\bibitem[Aleem et~al.(2024)Aleem, Wang, Maniparambil, Arazo, Dietlmeier, Curran, Connor, and Little]{aleem2024test}
Sidra Aleem, Fangyijie Wang, Mayug Maniparambil, Eric Arazo, Julia Dietlmeier, Kathleen Curran, Noel~EO' Connor, and Suzanne Little.
\newblock Test-time adaptation with salip: A cascade of sam and clip for zero-shot medical image segmentation.
\newblock In \emph{CVPR}, 2024.

\bibitem[Ali et~al.(2024)Ali, Essaid, Moalic, and Idoumghar]{ali2024review}
Muhammad~Junaid Ali, Mokhtar Essaid, Laurent Moalic, and Lhassane Idoumghar.
\newblock A review of automl optimization techniques for medical image applications.
\newblock \emph{Computerized Medical Imaging and Graphics}, 118:\penalty0 102441, 2024.

\bibitem[Bannur et~al.(2023)Bannur, Hyland, Liu, Perez-Garcia, Ilse, Castro, Boecking, Sharma, Bouzid, Thieme, et~al.]{bannur2023learning}
Shruthi Bannur, Stephanie Hyland, Qianchu Liu, Fernando Perez-Garcia, Maximilian Ilse, Daniel~C Castro, Benedikt Boecking, Harshita Sharma, Kenza Bouzid, Anja Thieme, et~al.
\newblock Learning to exploit temporal structure for biomedical vision-language processing.
\newblock In \emph{Proceedings of the IEEE/CVF Conference on Computer Vision and Pattern Recognition}, pp.\  15016--15027, 2023.

\bibitem[Bergstra et~al.(2011)Bergstra, Bardenet, Bengio, and K{\'e}gl]{bergstra2011tpe}
James Bergstra, R{\'e}mi Bardenet, Yoshua Bengio, and Bal{\'a}zs K{\'e}gl.
\newblock Algorithms for hyper-parameter optimization.
\newblock In \emph{NeurIPS}, 2011.

\bibitem[Bommasani et~al.(2021)Bommasani, Hudson, Adeli, et~al.]{Bommasani21FoundationModels}
Rishi Bommasani, Drew~A. Hudson, Ehsan Adeli, et~al.
\newblock On the opportunities and risks of foundation models.
\newblock \emph{arXiv preprint arXiv:2108.07258}, 2021.

\bibitem[Cassidy et~al.(2021)Cassidy, Kendrick, Reeves, Pappachan, O’Shea, Armstrong, and Yap]{cassidy2021diabetic}
Bill Cassidy, Connah Kendrick, Neil~D Reeves, Joseph~M Pappachan, Claire O’Shea, David~G Armstrong, and Moi~Hoon Yap.
\newblock Diabetic foot ulcer grand challenge 2021: evaluation and summary.
\newblock In \emph{Diabetic foot ulcers grand challenge}, pp.\  90--105. Springer, 2021.

\bibitem[Cheng et~al.(2023)Cheng, Ye, Deng, Chen, Li, Wang, Su, Huang, Chen, Jiang, et~al.]{cheng2023sam}
Junlong Cheng, Jin Ye, Zhongying Deng, Jianpin Chen, Tianbin Li, Haoyu Wang, Yanzhou Su, Ziyan Huang, Jilong Chen, Lei Jiang, et~al.
\newblock Sam-med2d.
\newblock \emph{arXiv preprint arXiv:2308.16184}, 2023.

\bibitem[Deng et~al.(2023)Deng, Li, He, et~al.]{Deng23SAMReviewMed}
Panyu Deng, Honghui Li, Jili He, et~al.
\newblock Segment anything model ({SAM}) for medical image segmentation: a survey.
\newblock \emph{arXiv preprint arXiv:2310.10617}, 2023.

\bibitem[Farina et~al.(2024)Farina, Franchi, Iacca, Mancini, and Ricci]{farina2024frustratingly}
Matteo Farina, Gianni Franchi, Giovanni Iacca, Massimiliano Mancini, and Elisa Ricci.
\newblock Frustratingly easy test-time adaptation of vision-language models.
\newblock \emph{Advances in Neural Information Processing Systems}, 37:\penalty0 129062--129093, 2024.

\bibitem[Gutman et~al.(2016)Gutman, Codella, Celebi, Helba, Marchetti, Mishra, and Halpern]{gutman2016skin}
David Gutman, Noel~CF Codella, Emre Celebi, Brian Helba, Michael Marchetti, Nabin Mishra, and Allan Halpern.
\newblock Skin lesion analysis toward melanoma detection: A challenge at the international symposium on biomedical imaging (isbi) 2016, hosted by the international skin imaging collaboration (isic).
\newblock \emph{arXiv preprint arXiv:1605.01397}, 2016.

\bibitem[He et~al.(2016)He, Zhang, Ren, and Sun]{resnet}
Kaiming He, Xiangyu Zhang, Shaoqing Ren, and Jian Sun.
\newblock Deep residual learning for image recognition.
\newblock In \emph{CVPR}, 2016.

\bibitem[He et~al.(2023)He, Bao, Gao, et~al.]{He23SAMMedical}
Sheng He, Runtian Bao, Jun Gao, et~al.
\newblock Accuracy of segment-anything model ({SAM}) in medical image segmentation: A comprehensive evaluation.
\newblock \emph{arXiv preprint arXiv:2304.09324}, 2023.

\bibitem[He et~al.(2021)He, Zhao, and Chu]{he2021automl}
Xin He, Kaiyong Zhao, and Xiaowen Chu.
\newblock Automl: A survey of the state-of-the-art.
\newblock \emph{Knowledge-based systems}, 212:\penalty0 106622, 2021.

\bibitem[Hesamian et~al.(2019)Hesamian, Jia, He, and Kennedy]{Hesamian19DeepReview}
Mohammad~Hesam Hesamian, Wenjing Jia, Xiangjian He, and Paul Kennedy.
\newblock Deep learning techniques for medical image segmentation: achievements and challenges.
\newblock \emph{JDI}, 2019.

\bibitem[Hoang et~al.(2024)Hoang, Vo, and Do]{hoang2024persistent}
Trung~Hieu Hoang, MinhDuc Vo, and Minh Do.
\newblock Persistent test-time adaptation in recurring testing scenarios.
\newblock \emph{Advances in Neural Information Processing Systems}, 37:\penalty0 123402--123442, 2024.

\bibitem[Iandola et~al.(2014)Iandola, Moskewicz, Karayev, Girshick, Darrell, and Keutzer]{densenet}
Forrest Iandola, Matt Moskewicz, Sergey Karayev, Ross Girshick, Trevor Darrell, and Kurt Keutzer.
\newblock Densenet: Implementing efficient convnet descriptor pyramids.
\newblock \emph{arXiv preprint arXiv:1404.1869}, 2014.

\bibitem[Isensee et~al.(2019)Isensee, Petersen, Klein, Zimmerer, Jaeger, Kohl, Wasserthal, Koehler, Norajitra, Wirkert, et~al.]{isensee2019nnu}
Fabian Isensee, Jens Petersen, Andre Klein, David Zimmerer, Paul~F Jaeger, Simon Kohl, Jakob Wasserthal, Gregor Koehler, Tobias Norajitra, Sebastian Wirkert, et~al.
\newblock nnu-net: Self-adapting framework for u-net-based medical image segmentation.
\newblock In \emph{Bildverarbeitung f{\"u}r die Medizin 2019: Algorithmen--Systeme--Anwendungen. Proceedings des Workshops vom 17. bis 19. M{\"a}rz 2019 in L{\"u}beck}, pp.\  22--22. Springer, 2019.

\bibitem[Jidney et~al.(2023)Jidney, Biswas, Abdullah~Al, Hossain, Alam, Talukder, Hossain, and Ullah]{jidney2023automl}
Tasmia~Tahmida Jidney, Angona Biswas, Nasim~Md Abdullah~Al, Ismail Hossain, Md~Jahangir Alam, Sajedul Talukder, Mofazzal Hossain, and Md~Azim Ullah.
\newblock Automl systems for medical imaging.
\newblock In \emph{Data driven approaches on medical imaging}, pp.\  91--106. Springer, 2023.

\bibitem[Kirillov et~al.(2023)Kirillov, Mintun, Ravi, Mao, Rolland, Gustafson, Xiao, Whitehead, Berg, Lo, Dollar, and Girshick]{Kirillov23SAM}
Alexander Kirillov, Eric Mintun, Nikhila Ravi, Hanzi Mao, Chloe Rolland, Laura Gustafson, Tete Xiao, Spencer Whitehead, Alexander~C. Berg, Wan-Yen Lo, Piotr Dollar, and Ross Girshick.
\newblock Segment anything.
\newblock In \emph{ICCV}, 2023.

\bibitem[Klinwichit et~al.(2023)Klinwichit, Yookwan, Limchareon, Chinnasarn, Jang, and Onuean]{klinwichit2023buu}
Podchara Klinwichit, Watcharaphong Yookwan, Sornsupha Limchareon, Krisana Chinnasarn, Jun-Su Jang, and Athita Onuean.
\newblock Buu-lspine: A thai open lumbar spine dataset for spondylolisthesis detection.
\newblock \emph{Applied Sciences}, 13\penalty0 (15):\penalty0 8646, 2023.

\bibitem[Koleilat et~al.(2024{\natexlab{a}})Koleilat, Asgariandehkordi, Rivaz, and Xiao]{koleilat2409medclip}
T~Koleilat, H~Asgariandehkordi, H~Rivaz, and Y~Xiao.
\newblock Medclip-samv2: Towards universal text-driven medical image segmentation. arxiv 2024.
\newblock \emph{arXiv preprint arXiv:2409.19483}, 2024{\natexlab{a}}.

\bibitem[Koleilat et~al.(2024{\natexlab{b}})Koleilat, Asgariandehkordi, Rivaz, and Xiao]{koleilat2024medclip}
Taha Koleilat, Hojat Asgariandehkordi, Hassan Rivaz, and Yiming Xiao.
\newblock Medclip-sam: Bridging text and image towards universal medical image segmentation.
\newblock In \emph{MICCAI}, 2024{\natexlab{b}}.

\bibitem[Kuş \& Aydin(2024)Kuş and Aydin]{Ku2024}
Zeki Kuş and Musa Aydin.
\newblock Medsegbench: A comprehensive benchmark for medical image segmentation in diverse data modalities.
\newblock \emph{Sci. Data}, 2024.

\bibitem[Litjens et~al.(2014)Litjens, Toth, Van De~Ven, Hoeks, Kerkstra, Van~Ginneken, Vincent, Guillard, Birbeck, Zhang, et~al.]{litjens2014evaluation}
Geert Litjens, Robert Toth, Wendy Van De~Ven, Caroline Hoeks, Sjoerd Kerkstra, Bram Van~Ginneken, Graham Vincent, Gwenael Guillard, Neil Birbeck, Jindang Zhang, et~al.
\newblock Evaluation of prostate segmentation algorithms for mri: the promise12 challenge.
\newblock \emph{MedIA}, 2014.

\bibitem[Litjens et~al.(2017)Litjens, Kooi, Bejnordi, Setio, Ciompi, Ghafoorian, Van Der~Laak, Van~Ginneken, and S{\'a}nchez]{litjens2017survey}
Geert Litjens, Thijs Kooi, Babak~Ehteshami Bejnordi, Arnaud Arindra~Adiyoso Setio, Francesco Ciompi, Mohsen Ghafoorian, Jeroen~Awm Van Der~Laak, Bram Van~Ginneken, and Clara~I S{\'a}nchez.
\newblock A survey on deep learning in medical image analysis.
\newblock \emph{MedIA}, 2017.

\bibitem[Liu et~al.(2024)Liu, Zeng, Ren, Li, Zhang, Yang, Jiang, Li, Yang, Su, Zhu, and Zhang]{liu2024groundingdino}
Shilong Liu, Zhaoyang Zeng, Tianhe Ren, Feng Li, Hao Zhang, Jie Yang, Qing Jiang, Chunyuan Li, Jianwei Yang, Hang Su, Jun Zhu, and Lei Zhang.
\newblock Grounding dino: Marrying dino with grounded pre-training for open-set object detection.
\newblock In \emph{Computer Vision – ECCV 2024: 18th European Conference, Milan, Italy, September 29–October 4, 2024, Proceedings, Part XLVII}, 2024.

\bibitem[Ma et~al.(2023)Ma, He, Li, et~al.]{Ma23MedSAM}
Jun Ma, Yuting He, Feifei Li, et~al.
\newblock {MedSAM}: Segment anything in medical images.
\newblock In \emph{MICCAI}, 2023.

\bibitem[Mazurowski et~al.(2023)Mazurowski, Dong, Gu, et~al.]{Mazurowski23SAMReview}
Maciej~A. Mazurowski, Haoyu Dong, Hanxue Gu, et~al.
\newblock Segment anything model for medical images?
\newblock \emph{arXiv preprint arXiv:2304.14660}, 2023.

\bibitem[Myronenko et~al.(2023)Myronenko, Yang, He, and Xu]{myronenko2023automated}
Andriy Myronenko, Dong Yang, Yufan He, and Daguang Xu.
\newblock Automated 3d segmentation of kidneys and tumors in miccai kits 2023 challenge.
\newblock In \emph{International Challenge on Kidney and Kidney Tumor Segmentation}, pp.\  1--7. Springer, 2023.

\bibitem[Oktay et~al.(2018)Oktay, Schlemper, Folgoc, et~al.]{Oktay18AttentionUnet}
Ozan Oktay, Jo~Schlemper, Loic~Le Folgoc, et~al.
\newblock Attention u-net: Learning where to look for the pancreas.
\newblock In \emph{MIDL}, 2018.

\bibitem[OpenAI(2023)]{chatgpt}
OpenAI.
\newblock Chatgpt (mar 14 version), 2023.
\newblock URL \url{https://chat.openai.com/chat}.

\bibitem[Oquab et~al.(2024)Oquab, Darcet, Moutakanni, Vo, Szafraniec, Khalidov, Fernandez, Haziza, Massa, El-Nouby, Assran, Ballas, Galuba, Howes, Huang, Li, Misra, Rabbat, Sharma, Synnaeve, Xu, Jegou, Mairal, Labatut, Joulin, and Bojanowski]{Oquab24DINOv2}
Maxime Oquab, Timoth{\'e}e Darcet, Th{\'e}o Moutakanni, Huy~V. Vo, Marc Szafraniec, Vasil Khalidov, Pierre Fernandez, Daniel Haziza, Francisco Massa, Alaaeldin El-Nouby, Mido Assran, Nicolas Ballas, Wojciech Galuba, Russell Howes, Po-Yao Huang, Shang-Wen Li, Ishan Misra, Michael Rabbat, Vasu Sharma, Gabriel Synnaeve, Hu~Xu, Herve Jegou, Julien Mairal, Patrick Labatut, Armand Joulin, and Piotr Bojanowski.
\newblock {DINO}v2: Learning robust visual features without supervision.
\newblock \emph{TMLR}, 2024.

\bibitem[Orlando et~al.(2020)Orlando, Fu, Breda, Van~Keer, Bathula, Diaz-Pinto, Fang, Heng, Kim, Lee, et~al.]{orlando2020refuge}
Jos{\'e}~Ignacio Orlando, Huazhu Fu, Jo{\~a}o~Barbosa Breda, Karel Van~Keer, Deepti~R Bathula, Andr{\'e}s Diaz-Pinto, Ruogu Fang, Pheng-Ann Heng, Jeyoung Kim, JoonHo Lee, et~al.
\newblock Refuge challenge: A unified framework for evaluating automated methods for glaucoma assessment from fundus photographs.
\newblock \emph{MedIA}, 2020.

\bibitem[Papandreou et~al.(2015)Papandreou, Chen, Murphy, and Yuille]{Papandreou15Weakly}
George Papandreou, Liang-Chieh Chen, Kevin~P. Murphy, and Alan~L. Yuille.
\newblock Weakly-and semi-supervised learning of a deep convolutional network for semantic image segmentation.
\newblock In \emph{ICCV}, 2015.

\bibitem[Patil \& Deore(2013)Patil and Deore]{patil2013medical}
Dinesh~D Patil and Sonal~G Deore.
\newblock Medical image segmentation: a review.
\newblock \emph{IJCSMC}, 2013.

\bibitem[Pogorelov et~al.(2017)Pogorelov, Randel, Griwodz, Eskeland, de~Lange, Johansen, Spampinato, Dang-Nguyen, Lux, Schmidt, Riegler, and Halvorsen]{Pogorelov:2017:KMI:3083187.3083212}
Konstantin Pogorelov, Kristin~Ranheim Randel, Carsten Griwodz, Sigrun~Losada Eskeland, Thomas de~Lange, Dag Johansen, Concetto Spampinato, Duc-Tien Dang-Nguyen, Mathias Lux, Peter~Thelin Schmidt, Michael Riegler, and P{\aa}l Halvorsen.
\newblock Kvasir: A multi-class image dataset for computer aided gastrointestinal disease detection.
\newblock In \emph{ACM MMSys}, 2017.

\bibitem[Ronneberger et~al.(2015)Ronneberger, Fischer, and Brox]{Ronneberger15Unet}
Olaf Ronneberger, Philipp Fischer, and Thomas Brox.
\newblock U-net: Convolutional networks for biomedical image segmentation.
\newblock In \emph{MICCAI}, 2015.

\bibitem[Sandler et~al.(2018)Sandler, Howard, Zhu, Zhmoginov, and Chen]{mobilenetv2}
Mark Sandler, Andrew Howard, Menglong Zhu, Andrey Zhmoginov, and Liang-Chieh Chen.
\newblock Mobilenetv2: Inverted residuals and linear bottlenecks.
\newblock In \emph{CVPR}, 2018.

\bibitem[Shin \& Kim(2024)Shin and Kim]{shin2024tta}
Jin Shin and Hyun Kim.
\newblock L-tta: Lightweight test-time adaptation using a versatile stem layer.
\newblock \emph{Advances in Neural Information Processing Systems}, 37:\penalty0 39325--39349, 2024.

\bibitem[siatsyx(2024)]{ct2usforkidneyseg2024}
siatsyx.
\newblock {CT2USforKidneySeg} -- kidney segmentation from ct to ultrasound transfer.
\newblock \url{https://www.kaggle.com/datasets/siatsyx/ct2usforkidneyseg/data}, 2024.
\newblock Accessed: 2024-07-08.

\bibitem[Snell et~al.(2017)Snell, Swersky, and Zemel]{Snell17Prototypical}
Jake Snell, Kevin Swersky, and Richard Zemel.
\newblock Prototypical networks for few-shot learning.
\newblock In \emph{NeurIPS}, 2017.

\bibitem[Snoek et~al.(2012)Snoek, Larochelle, and Adams]{snoek2012practical}
Jasper Snoek, Hugo Larochelle, and Ryan~P Adams.
\newblock Practical bayesian optimization of machine learning algorithms.
\newblock In \emph{NeurIPS}, 2012.

\bibitem[Song et~al.(2022)Song, Zheng, Lei, Ni, Zhao, and Hu]{song2022ct2us}
Yuxin Song, Jing Zheng, Long Lei, Zhipeng Ni, Baoliang Zhao, and Ying Hu.
\newblock Ct2us: Cross-modal transfer learning for kidney segmentation in ultrasound images with synthesized data.
\newblock \emph{Ultrasonics}, 2022.

\bibitem[Sun et~al.(2020)Sun, Wang, Liu, Miller, Efros, and Hardt]{Sun2020TTT}
Yu~Sun, Xiaolong Wang, Zhuang Liu, John Miller, Alexei Efros, and Moritz Hardt.
\newblock Test-time training with self-supervision for generalization under distribution shifts.
\newblock In \emph{ICML}, 2020.

\bibitem[Tan \& Le(2019)Tan and Le]{efficientnet}
Mingxing Tan and Quoc Le.
\newblock Efficientnet: Rethinking model scaling for convolutional neural networks.
\newblock In \emph{ICML}, 2019.

\bibitem[Tang et~al.(2024)Tang, Jiang, Shen, Zhang, Chen, and Li]{tang2024chain}
Lv~Tang, Peng-Tao Jiang, Zhi-Hao Shen, Hao Zhang, Jin-Wei Chen, and Bo~Li.
\newblock Chain of visual perception: Harnessing multimodal large language models for zero-shot camouflaged object detection.
\newblock In \emph{ACM-MM}, 2024.

\bibitem[{Vision and Image Processing Lab, University of Waterloo}(2021)]{uw_skin_cancer_2021}
{Vision and Image Processing Lab, University of Waterloo}.
\newblock Skin cancer database.
\newblock \url{https://uwaterloo.ca/vision-image-processing-lab/research-demos/skin-cancer-detection}, 2021.

\bibitem[Wang et~al.(2020)Wang, Shelhamer, Liu, Olshausen, and Darrell]{Wang2020Tent}
Dequan Wang, Evan Shelhamer, Shaoteng Liu, Bruno Olshausen, and Trevor Darrell.
\newblock Tent: Fully test-time adaptation by entropy minimization.
\newblock \emph{arXiv preprint arXiv:2006.10726}, 2020.

\bibitem[Wang et~al.(2024)Wang, Lv, Yu, Hong, Qi, Wang, Ji, Yang, Zhao, XiXuan, et~al.]{wang2024cogvlm}
Weihan Wang, Qingsong Lv, Wenmeng Yu, Wenyi Hong, Ji~Qi, Yan Wang, Junhui Ji, Zhuoyi Yang, Lei Zhao, Song XiXuan, et~al.
\newblock Cogvlm: Visual expert for pretrained language models.
\newblock \emph{NeurIPS}, 2024.

\bibitem[Wu \& Xu(2024)Wu and Xu]{wu2024one}
Junde Wu and Min Xu.
\newblock One-prompt to segment all medical images.
\newblock In \emph{CVPR}, 2024.

\bibitem[Xian et~al.(2019)Xian, Lampert, Schiele, and Akata]{Xian19ZSLSurvey}
Yongqin Xian, Christoph~H. Lampert, Bernt Schiele, and Zeynep Akata.
\newblock Zero-shot learning---a comprehensive evaluation of the state of the art.
\newblock \emph{IEEE TPAMI}, 2019.

\bibitem[Xu et~al.(2023)Xu, Zhang, Wei, Hu, and Bai]{xu2023side}
Mengde Xu, Zheng Zhang, Fangyun Wei, Han Hu, and Xiang Bai.
\newblock Side adapter network for open-vocabulary semantic segmentation.
\newblock In \emph{CVPR}, 2023.

\bibitem[Yang et~al.(2024)Yang, Bi, Zhang, and Sun]{yang2024sam}
Sihan Yang, Haixia Bi, Hai Zhang, and Jian Sun.
\newblock Sam-unet: Enhancing zero-shot segmentation of sam for universal medical images.
\newblock \emph{arXiv preprint arXiv:2408.09886}, 2024.

\bibitem[Yu et~al.(2023{\natexlab{a}})Yu, Wang, Tang, Feng, and Lv]{yu2023eu}
Caiyang Yu, Yixi Wang, Chenwei Tang, Wentao Feng, and Jiancheng Lv.
\newblock Eu-net: Automatic u-net neural architecture search with differential evolutionary algorithm for medical image segmentation.
\newblock \emph{Computers in Biology and Medicine}, 167:\penalty0 107579, 2023{\natexlab{a}}.

\bibitem[Yu et~al.(2023{\natexlab{b}})Yu, Wang, Zhao, and Gao]{mixvit}
Xiaohan Yu, Jun Wang, Yang Zhao, and Yongsheng Gao.
\newblock Mix-vit: Mixing attentive vision transformer for ultra-fine-grained visual categorization.
\newblock \emph{Pattern Recognit.}, 2023{\natexlab{b}}.

\bibitem[Zhang et~al.(2024)Zhang, Xu, Usuyama, Xu, Bagga, Tinn, Preston, Rao, Wei, Valluri, Wong, Tupini, Wang, Mazzola, Shukla, Liden, Gao, Crabtree, Piening, Bifulco, Lungren, Naumann, Wang, and Poon]{zhang2024biomedclip}
Sheng Zhang, Yanbo Xu, Naoto Usuyama, Hanwen Xu, Jaspreet Bagga, Robert Tinn, Sam Preston, Rajesh Rao, Mu~Wei, Naveen Valluri, Cliff Wong, Andrea Tupini, Yu~Wang, Matt Mazzola, Swadheen Shukla, Lars Liden, Jianfeng Gao, Angela Crabtree, Brian Piening, Carlo Bifulco, Matthew~P. Lungren, Tristan Naumann, Sheng Wang, and Hoifung Poon.
\newblock A multimodal biomedical foundation model trained from fifteen million image–text pairs.
\newblock \emph{NEJM AI}, 2024.

\bibitem[Zhao et~al.(2025)Zhao, Gu, Yang, Usuyama, Lee, Kiblawi, Naumann, Gao, Crabtree, Abel, et~al.]{zhao2025foundation}
Theodore Zhao, Yu~Gu, Jianwei Yang, Naoto Usuyama, Ho~Hin Lee, Sid Kiblawi, Tristan Naumann, Jianfeng Gao, Angela Crabtree, Jacob Abel, et~al.
\newblock A foundation model for joint segmentation, detection and recognition of biomedical objects across nine modalities.
\newblock \emph{Nature methods}, 2025.

\bibitem[Zhao et~al.(2023)Zhao, Zhang, Wu, Zhang, Zhang, Wang, and Xie]{zhao2023one}
Ziheng Zhao, Yao Zhang, Chaoyi Wu, Xiaoman Zhang, Ya~Zhang, Yanfeng Wang, and Weidi Xie.
\newblock One model to rule them all: Towards universal segmentation for medical images with text prompts.
\newblock \emph{CoRR}, 2023.

\bibitem[Zhou et~al.(2018)Zhou, Siddiquee, Tajbakhsh, and Liang]{Zhou18UnetPlusPlus}
Zongwei Zhou, Md~Mahfuzur~Rahman Siddiquee, Nima Tajbakhsh, and Jianming Liang.
\newblock Unet++: A nested u-net architecture for medical image segmentation.
\newblock In \emph{DLMIA}, 2018.

\bibitem[Zuiderveld et~al.(1994)]{zuiderveld1994contrast}
Karel~J Zuiderveld et~al.
\newblock Contrast limited adaptive histogram equalization.
\newblock \emph{Graphics gems}, 1994.

\end{thebibliography}
\bibliographystyle{iclr2026_conference}

\newpage
\appendix
\section{Visual Grounding Models}

Visual Grounding (also known as Referring Expression Comprehension) is a crucial task in
computer vision and natural language processing that aims to localize specific objects or
regions in an image based on a natural language textual description (a "text prompt" or
"referring expression").
Unlike traditional object detection systems that identify objects from a predefined set of
categories, visual grounding models are designed to understand free-form, often complex,
textual prompts that can describe an object by its attributes, spatial relationships to
other objects, or other distinguishing characteristics.

The input to a typical visual grounding model consists of an image and a text prompt
indicating the object of interest.
The output is usually the set of coordinates for a bounding box that precisely encloses
the described object within the image.
This capability to "ground" language in visual content makes these models powerful tools
for fine-grained image understanding and interaction.

The ability of these models to interpret nuanced language and map it to specific image
regions makes them highly valuable for various downstream applications, including
human-computer interaction, robotics, image editing, and, as in our work, providing
precise visual prompts (bounding boxes) for subsequent processing stages like segmentation.
Given their relatively recent surge in capability and application, a brief introduction is
warranted for reviewers who may be less familiar with this specific class of models.

For our training-free medical image segmentation pipeline, we required a robust and
versatile visual grounding model capable of accurately localizing objects of interest based
on textual prompts, which would affect the location and quality of masks generated by the successory mask proposing modules.

We selected \textbf{CogVLM} \citep{wang2024cogvlm} as the grounding module in our
pipeline for several key reasons:

\begin{enumerate}
    \item \textbf{State-of-the-Art Performance and Open Access:} CogVLM is a powerful,
    open-source visual language model that has demonstrated strong performance across a
    wide array of vision-language tasks, including those requiring fine-grained
    understanding and localization. Its availability facilitates reproducibility and
    further research.

    \item \textbf{Effective Grounding Capabilities:} While CogVLM is a general visual
    language model, its architecture is inherently well-suited for tasks that require
    grounding textual concepts in images. It can be effectively prompted or adapted to
    output bounding box coordinates corresponding to objects described in the text. Its
    ability to handle detailed descriptions and disambiguate objects makes it suitable
    for generating precise visual prompts.

    \item \textbf{Strong Generalization and Zero-Shot Potential:} Due to its extensive
    pre-training on large-scale image-text datasets, CogVLM exhibits impressive
    generalization capabilities to novel objects and scenarios. This strong zero-shot
    or few-shot performance aligns perfectly with the "training-free" philosophy of our
    proposed pipeline, allowing us to leverage its capabilities without task-specific
    fine-tuning for the grounding step.

\end{enumerate}

% In summary, CogVLM's combination of open-source accessibility, advanced architecture,
% strong performance in visual-language understanding, and excellent generalization
% capabilities made it an ideal choice for the text-to-box grounding module in our proposed
% training-free segmentation framework.

\section{Visual Prompt Boosting}
\label{supp:prmptAug}

While the bounding box $B$ provides a coarse localization, promptable segmentation models often benefit from more precise internal points to disambiguate the target object from the background or adjacent structures. We generate these point prompts automatically by analyzing the visual features within the image $I$. This process is deterministic given the chosen pre-trained encoder.

\paragraph{Anchor Point Selection.} We define an initial anchor point $p_a = (x_a, y_a)$ as the geometric center of the bounding box $B$.
\begin{equation}
    x_a = \frac{x_{min} + x_{max}}{2}, \quad y_a = \frac{y_{min} + y_{max}}{2}
    \label{eq:anchor_point}
\end{equation}

\paragraph{Dense Feature Extraction.} We employ a pre-trained vision encoder, specifically DINOv2 \citep{Oquab24DINOv2} known for its strong semantic feature representation capabilities without fine-tuning, to extract dense feature maps $F \in \mathbb{R}^{H' \times W' \times D}$ from the image $I$. Let $f(p) \in \mathbb{R}^D$ denote the feature vector corresponding to a spatial location $p=(x,y)$ in the image (potentially requiring interpolation if $p$ does not align perfectly with the feature grid). Let $f_a = f(p_a)$ be the feature vector of the anchor point.

\paragraph{Similarity-Based Point Selection.} We identify candidate points within the image (or potentially restricted to the bounding box $B$ for efficiency) that are semantically similar to the anchor point in the DINOv2 feature space. We compute the cosine similarity between the anchor feature $f_a$ and all other features $f(p)$ and select the top-$k$ points $P_k = \{p_1, ..., p_k\}$ with the highest similarity scores:
\begin{equation}
    P_k = \underset{p \in I, p \neq p_a}{\text{top-}k} \left( \frac{f_a \cdot f(p)}{\|f_a\| \|f(p)\|} \right)
    \label{eq:topk_points}
\end{equation}
In our experiments, we typically use $k=10$.

\paragraph{Point Clustering and Center Calculation.} The top-$k$ similar points might form distinct spatial clusters within the target object. To obtain representative points covering potentially different parts of the object, we cluster the coordinates of the points in $P_k$ into $n$ groups using K-Means clustering.
\begin{equation}
    C_1, ..., C_n = \kmeans( \{ (x_i, y_i) \mid p_i \in P_k \}, n )
    \label{eq:kmeans}
\end{equation}
where $C_j$ represents the set of points belonging to the $j$-th cluster. We set $n=3$ in our typical configuration.

\paragraph{Final Point Prompt Generation.} We calculate the centroid (mean coordinate) $p_{c,j}$ for each cluster $C_j$. These centroids form our set of additional positive point prompts $P_c = \{p_{c,1}, ..., p_{c,n}\}$. The final set of visual prompts for the segmentation model consists of the bounding box $B$ and the generated points $P_c$.
\section{Learnable Test-time Adaptors}

We introduced a set of domain-adapted image transformations within our Learnable Test-time Adaptors (LTAs) in an effort to reduce the domain shift between clinical images and foundation models pretrained on natural images. We applied a series of light-weighted image transformation and prompt enhancement during the test time. The operations are designed to enhance local contrast, simulate modality-specific variability and emphasize the anatomical features that are hard to detect within the original images. 

All the transformations have a couple of tunable parameters, and we optimize over them using Bayesian optimization (TPE~\citep{bergstra2011tpe}). We apply these transformations separately to the grounding input and segmentation input, each with their own set of parameters. We also include two non-augmentation parameters in the search: the grounding prompt index $i_{grd}$  controls which predefined prompt to use for the grounding model, and the number of enhanced prompt points $k$ controls the number of point clusters used to boost the box prompt. These parameters affect the quality and stability of the grounding and segmentation process, especially in noisy or ambiguous settings. A complete description of the involved operations is as follows.

\begin{itemize}
\item \textbf{HSV Shift:} Shift of image hue, saturation, and brightness. This captures variations in scanner settings or lighting.
\item \textbf{RGB Shift:} Adds individual offsets to all of the R/G/B channels. This is used to reduce the model's dependence on an even intensity distribution.
\item \textbf{CLAHE~\citep{zuiderveld1994contrast}:} Performs contrast-limited adaptive histogram equalization for greater visibility in locally low-contrast areas.
\item \textbf{Unsharp Masking:} Sharpens edges by subtracting a blurred version of the image, making anatomical boundaries clearer.
\item \textbf{Grounding Prompt Selection:} Controls the choice of a predefined bounding box prompt used for the grounding model. Different prompt configurations can significantly affect alignment between the language and the vision encoder.
\item \textbf{Number of Boosted Points:} Controls the number of points used for box prompt boosting. The choice can be 0, which means prompt boosting is not performed. In practice, we find this component is probably learned to be skipped if the target region contains diverse visual patterns and the central point is not very representative. Otherwise, the prompt boosting module tends to generate a few representative points as additional prompts and make the segmentation boundary more accurate. 
\end{itemize}

We list the hyperparameters and their search spaces below in Table~\ref{tab:transformation_hp}. These parameters are optimized using Bayesian optimization with a Tree-structured Parzen Estimator (TPE), maximizing proxy validation scores, and were chosen to provide a moderate set of variations without over-warping the input.

\begin{table}[t]
\caption{Hyperparameters and search spaces used in LTA optimization.}
\label{tab:transformation_hp}
\centering
\small
\renewcommand{\arraystretch}{1.1}
\adjustbox{max width=\textwidth}{
\begin{tabular}{lllll}
\toprule
\textbf{Operation} & \textbf{Hyperparameter} & \textbf{Type} & \textbf{Range} & \textbf{Description} \\
\midrule
\multirow{3}{*}{HSV Shift} 
& hsv\_hue\_shift & Integer & [0, 20] & Amount of hue rotation \\
& hsv\_sat\_shift & Integer & [0, 30] & Change in color saturation \\
& hsv\_val\_shift & Integer & [0, 30] & Change in image brightness \\
\midrule
\multirow{3}{*}{RGB Shift} 
& r\_shift & Integer & [0, 20] & Red channel offset \\
& g\_shift & Integer & [0, 20] & Green channel offset \\
& b\_shift & Integer & [0, 20] & Blue channel offset \\
\midrule
\multirow{2}{*}{CLAHE}     
& clahe\_clip & Float   & [0.0, 4.0] & Clip limit for local contrast \\
& clahe\_grid & Integer & [1, 4]     & Number of tiles per image axis \\
\midrule
Unsharp Masking            
& edge\_strength & Float & [0.0, 1.0] & Edge enhancement level \\
\midrule
Prompt Selection 
& grd\_prompt\_id & Categorical & \{0,1,\dots,9\} & Choice of grounding prompt ID \\
\midrule
Prompt Boosting 
& bst\_k\_points & Integer & [0, 5] & Number of point prompts \\
\bottomrule
\end{tabular}
}
\end{table}

%These transformations are selectively applied to the grounding input ($P_{\text{grd}}$) and the segmentation input ($P_{\text{seg}}$) with separate parameters. During optimization, we independently search for both sets to best align the intermediate representations with the domain-specific characteristics of the target image. %The ablation results in Table~\ref{tab:performance_comparison} of the main document demonstrate the effectiveness of this decoupled transformation strategy.

\section{Pre-defined LLM Templates}
We use the following templates to generate necessary text resources used in our automatic medical image segmentation pipeline. 

\subsection{Text Prompt for the Grounding Model}

\begin{tcolorbox}[title=Input to ChatGPT-4o]
I want to describe the visual characteristics of the $\textcolor{red}{T_{target}}$ in an image of $\textcolor{blue}{T_{whole}}$. Please generate 10 sentences for the purpose of computer vision annotation. Each sentence should describe the shape, size (relative to the entire image), color, texture, and other visual features that help humans easily recognize the target. Each sentence must begin with "Locate the  $\textcolor{red}{T_{target}}$ in the image of $\textcolor{blue}{T_{whole}}$, which". Each sentence should be limited within 20 words. Ensure the 10 sentences are diverse and distinct. Write the results in a .txt file, with each description on a separate line and without line numbers.
\end{tcolorbox}

\begin{tcolorbox}[title=Sample Output for the REFUGE dataset]
\scriptsize
Locate the optical disc in the image of eye fundus, which is round, bright, and positioned toward the nasal side.\\
Locate the optical disc in the image of eye fundus, which appears yellowish, slightly elevated, and sharply bordered by retina.\\
Locate the optical disc in the image of eye fundus, which shows a pale circular region where blood vessels radiate outward.\\
Locate the optical disc in the image of eye fundus, which is circular, bright pink, and textured with fine vascular structures.\\
Locate the optical disc in the image of eye fundus, which looks like a smooth light-colored oval near the nasal margin.\\
Locate the optical disc in the image of eye fundus, which appears as a distinct white spot with central vessel branching.\\
Locate the optical disc in the image of eye fundus, which displays a soft granular texture and contrasts with darker retina.\\
Locate the optical disc in the image of eye fundus, which looks bright, circular, and located slightly off-center to the left.\\
Locate the optical disc in the image of eye fundus, which appears as a light area with radial vascular convergence.\\
Locate the optical disc in the image of eye fundus, which is smooth, round, pale, and lies near the nasal boundary.
\end{tcolorbox}

\subsection{Generating Contrastive Classes for the Validation Model}
\begin{tcolorbox}[title=Input to ChatGPT-4o]
In an image of $\textcolor{blue}{T_{whole}}$, what are the typical regions that have visual features and positions distinct from the $\textcolor{red}{T_{target}}$? List 2 to 4 of the most visually different parts or regions, using names commonly found in medical literature. Write the results in a .txt file, with the names separated by comma without additional output. 
\end{tcolorbox}

\begin{tcolorbox}[title=Sample Output for the REFUGE dataset]
macula, fovea, retinal blood vessels, peripheral retina
\end{tcolorbox}

In our pipeline, we manually add "background" as a default contrastive class for each dataset in addition to the generated classes. We empirically find this is beneficial to validate the foreground targets with the visual language model $\mathrm{M}_{val}$. 

\subsection{Generating Target Descriptions for the Validation Model}
We use a similar template as that used for the grounding model, only differing in the requirements of prefix and length of results. 
\begin{tcolorbox}[title=Input to ChatGPT-4o]
I want to describe the visual characteristics of the $\textcolor{red}{T_{target}}$ in an image of $\textcolor{blue}{T_{whole}}$. Please generate 10 sentences for the purpose of computer vision annotation. Each sentence should describe the shape, size (relative to the entire image), color, texture, and other visual features that help humans easily recognize the target. Each sentence must begin with "It appears", "It looks" or "It is".  Ensure the 10 sentences are diverse and distinct. Write the results in a .txt file, with each description on a separate line and without line numbers.
\end{tcolorbox}

\begin{tcolorbox}[title=Sample Output for the REFUGE dataset]
\footnotesize
It appears as a round, pale yellow region slightly off-center, with a smooth texture and well-defined margins that stand out against the darker retina. \\
It looks like a circular, elevated area occupying about 10\% of the image width, with a pinkish hue and radiating blood vessels. \\
It is an oval-shaped, bright region located nasally, with a creamy white color and a central depression known as the optic cup. \\
It appears as a clearly demarcated disk with a light peach tone, fine granular texture, and a ring-like structure due to the cup-to-disc contour. \\
It looks like a small, reflective area with a yellow-white coloration, prominent vessel branches emerging from its center, and a slightly rough surface. \\
It is a slightly oval, pink-colored disk occupying approximately one-eighth of the image, with a smooth edge and a central dip. \\
It appears as a bright, circular area with a distinct boundary, a pale central cup, and a surrounding rim that contrasts with the adjacent retina. \\
It looks like a sunken, light-toned region, slightly raised at the edges, and serves as the entry and exit point for retinal blood vessels. \\
It is a well-circumscribed, yellowish-white disk, relatively small compared to the full image, with a visible vascular pattern originating from its core. \\
It appears as a doughnut-like structure with a pink outer ring and a paler inner cup, located toward the nasal side and distinct in texture from the rest of the fundus.
\end{tcolorbox}
\newpage

\section{More Qualitative Examples}
\begin{figure}[htbp]
    \centering
    \includegraphics[width=\textwidth]{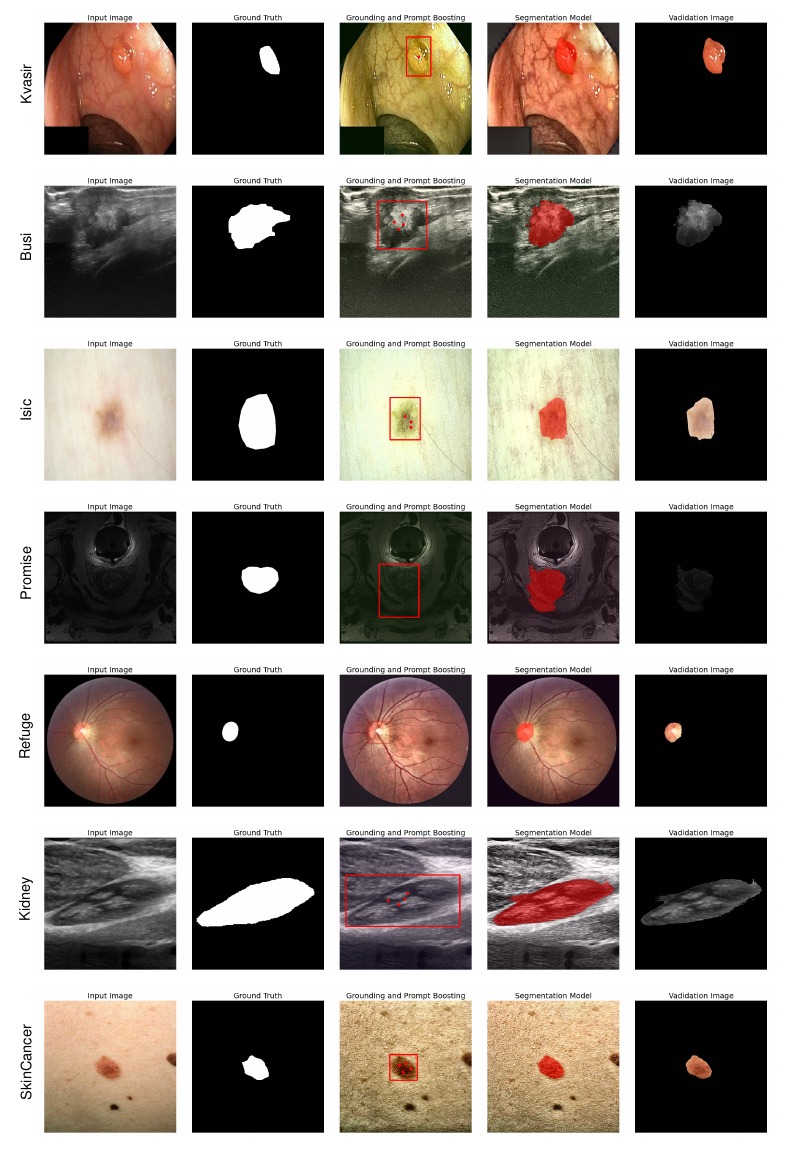}
    \caption{More qualitative results with our automatic segmentation pipeline.}
    \label{fig:showcase_more}
\end{figure}

\newpage
\section{Comparison with text-prompted supervised models}

The recent state-of-the-art medical image segmentation model BiomedParse~\citep{zhao2025foundation} also accepts text prompts at inference. However, it is not a genuine zero-shot learning model. Its text-conditioning ability is learned in a supervised manner, where the textual inputs during training are largely restricted to domain-specific terms and their limited variations contained in the annotated datasets. Consequently, its zero-shot capability does not extend to unseen domains or novel terminology, and the model exhibits poor generalization on out-of-domain datasets, being inherently constrained by the supervised bias of its training data.

To more comprehensively assess the robustness of our approach and such baselines, we additionally compare performance on datasets outside the domain of BiomedParse \citep{zhao2025foundation}. The datasets used are the BUU (Lumbar Vertebrae X-ray) dataset \citep{klinwichit2023buu} and the FUSC2021 (Diabetic Foot Ulcer) dataset \citep{cassidy2021diabetic} .  These results have been presented in Table \ref{tab:ood_performance}. On datasets included in BiomedParse’s training data, BiomedParse achieves performance comparable to the evaluated supervised models (as observed in Table \ref{tab:performance_on_datasets}), as expected. However, on unseen datasets, our approach substantially outperforms BiomedParse as shown in Table \ref{tab:ood_performance}.

\begin{table}[h]
\caption{Segmentation performance comparison across out-of-domain datasets between BiomedParse and our approach.}
\label{tab:ood_performance}
\centering
\small
\renewcommand{\arraystretch}{1.1}
\adjustbox{max width=\textwidth}{
\begin{tabular}{lccc}
\toprule
Method & BUU (Lumbar Vertebrae X-ray) & FUSC2021 (Diabetic Foot Ulcer) \\
\midrule
BiomedParse \citep{zhao2025foundation} & 4.25	& 21.61 \\
AutoMiSeg \textbf{(Ours)}  & 67.24	& 56.19 \\
\bottomrule
\end{tabular}
}
\end{table}

\section{Inference Time Comparison with Other Zero-Shot Methods}

Here we compare the inference times of the different zero-shot segmentation methods with our approach.  All inferences were run on an Nvidia A100 GPU, and the average per-sample inference time (in seconds) over 100 samples from the Busi dataset was computed. These values are presented in Table \ref{tab:inference_times}.

Although our pipeline does introduce more components to address the challenging zero-shot problem in medical domains, many of these components are only involved in the one-time adaptation process. Only the grounding and segmentation models are used during inference.  As seen in Table \ref{tab:inference_times}, our approach achieves a comparable inference speed to the baselines.

\begin{table}[h]
\caption{Comparison of average per-sample inference times over 100 samples from the Busi \citep{walid_al-dhabyani_mohammed_gomaa_hussien_khaled_aly_fahmy_2024} dataset between other zero-shot segmentation baselines and our approach. All values are in seconds.}
\label{tab:inference_times}
\centering
\small
\renewcommand{\arraystretch}{1.1}
\adjustbox{max width=\textwidth}{
\begin{tabular}{cccc}
\toprule
SaLIP \citep{aleem2024test} & MedCLIP-SAM \citep{koleilat2024medclip} & MedCLIP-SAM-v2 \citep{koleilat2409medclip} & AutoMiSeg \textbf{(Ours)} \\
\midrule
1.50 & 0.49	& 2.09 & 1.56 \\
\bottomrule
\end{tabular}
}
\end{table}

\section{Modular Ablation Study}
We perform an ablation study to evaluate our choice of choosing the base variant of CogVLM as our grounding model for target area grounding. We replace CogVLM with the base variant of Grounding Dino \citep{liu2024groundingdino}, while keeping all other components of the pipeline fixed.  We evaluate the performance on the Kvasir and Busi datasets, using the same grounding prompts that were used with CogVLM. Since Grounding Dino outputs a number of bounding boxes along with their scores, we select the bounding box with the top score. Similarly, we also evaluate suitable variants for the segmentation and validator modules. For segmentation, we replace the default SAM with MedSAM and SAM-Med2D. For the validator, we replace the default BiomedCLIP with BioVIL-T \citep{bannur2023learning}. The results of the ablation study are presented in Table \ref{tab:grounding_ablation}.

\begin{table}[H]
\caption{Segmentation performance under ablation when evaluating CogVLM and Grounding DINO separately as grounding models on Kvasir \citep{Pogorelov:2017:KMI:3083187.3083212} and Busi \citep{walid_al-dhabyani_mohammed_gomaa_hussien_khaled_aly_fahmy_2024} datasets.}
\label{tab:grounding_ablation}
\centering
\small
\renewcommand{\arraystretch}{1.1}
\adjustbox{max width=\textwidth}{
\begin{tabular}{llcc}
\toprule
\text{Module} & \text{Ablation} & Kvasir  (endoscopy) & 
Busi (ultrasound) \\
\midrule
- & AutoMiSeg default & 74.80 & 61.65 \\
\midrule
Grounding & CogVLM $\rightarrow$ Grounding Dino & 67.80 & 51.50 \\
\midrule
\multirow{2}{*}{Segmentation} & SAM $\rightarrow$ MedSAM & 78.50 & 66.81 \\
& SAM $\rightarrow$ SAM-Med2D & 72.88 & 66.30 \\
\midrule
Validator & BiomedCLIP $\rightarrow$ BioVIL-T & 73.60 & 60.65 \\
\bottomrule
\end{tabular}
}
\end{table}

\end{document}